\documentclass{article}
\usepackage{iclr2027_conference,times}
\iclrfinalcopy
\usepackage{amsmath,amsfonts,bm}

\def\eqref#1{equation~\ref{#1}}

\def\1{\bm{1}}

\DeclareMathAlphabet{\mathsfit}{\encodingdefault}{\sfdefault}{m}{sl}
\SetMathAlphabet{\mathsfit}{bold}{\encodingdefault}{\sfdefault}{bx}{n}

\usepackage{amsmath}
\usepackage{amssymb}
\usepackage{booktabs}
\usepackage{graphicx}
\usepackage{float}
\usepackage{microtype}
\usepackage{hyperref}
\usepackage{url}
\usepackage{longtable}
\usepackage{placeins}
\usepackage{needspace}
\usepackage{titletoc}
\usepackage[most]{tcolorbox}
\usepackage{listings}

\definecolor{appendixgreen}{HTML}{27815C}
\definecolor{appendixlight}{HTML}{F7FBF8}
\newcounter{appendixlisting}
\newtcblisting{agentlisting}[1]{
  enhanced,breakable,listing only,
  colback=appendixlight,colframe=appendixgreen,
  colbacktitle=appendixgreen,coltitle=white,
  fonttitle=\small\bfseries,boxrule=0.4pt,arc=1.5pt,
  left=5pt,right=5pt,top=4pt,bottom=4pt,
  before skip=8pt,after skip=8pt,
  title={\refstepcounter{appendixlisting}Listing~\theappendixlisting: #1},
  listing options={basicstyle=\ttfamily\footnotesize,breaklines=true,
    columns=fullflexible,keepspaces=true,showstringspaces=false}
}

\definecolor{colorLawrence}{RGB}{200,0,0}
\definecolor{colorPranjal}{RGB}{72,61,139}
\definecolor{colorJY}{RGB}{255,165,0}

\def\BenchmarkName{\texttt{cua-speedrun}}

\title{\BenchmarkName{}: Standardized Benchmarking of the Speed of Computer-Use Agents}

\author{%
  \textbf{Pranjal Aggarwal}\thanks{Equal contribution.} \quad
  \textbf{Lawrence Keunho Jang}\footnotemark[1] \quad
  \textbf{Sean Welleck} \quad
  \textbf{Daniel Fried} \\
  \textbf{Ruslan Salakhutdinov} \quad
  \textbf{Jing Yu Koh}\footnotemark[1] \\[0.5em]
  {\normalfont Carnegie Mellon University} \\
  {\normalfont\texttt{\{pranjala,ljang,jingyuk\}@cs.cmu.edu}}
}

\begin{document}
\maketitle
\lhead{Preprint}
\begin{abstract}
Computer use agents (CUAs), which use graphical user interfaces (GUIs) to complete tasks on a computer, have recently surpassed human performance on many standard benchmarks, including difficult long-horizon tasks. Their capabilities are undoubtedly impressive,
however, a key barrier to the widespread adoption and deployment of CUAs remains their speed and cost.
Progress towards faster yet capable CUAs requires reliable evaluation of their speed, but many CUA benchmarks currently face a reproducibility crisis.
Benchmarks are based on complex infrastructure with varying machine and container configurations that confound the evaluation of the execution speed of CUAs. Towards addressing this gap, we propose \BenchmarkName{}, which introduces standardized infrastructure and task sets, with a focus on evaluating the speed and efficiency of CUAs. 
\BenchmarkName{} uses a uniform virtual machine setup and execution pipeline, along with a common agent interface that enables single-agent implementations to operate seamlessly across different benchmarks. 
Across four different CUA benchmarks, we evaluate how reasoning effort, agent harnesses, and environment latency affect performance, speed, and cost. We find no single model family is optimal for all three; none of the open-weight models are on the frontier, and also, unintuitively, for some models increasing the reasoning effort can speed up task completion, while faster environment input-output can slow down overall task completion time. We also demonstrate that we can effectively reduce the evaluation task set of most CUA benchmarks without degrading overall statistical power, allowing for more efficient benchmarking and comparison. 
We believe \BenchmarkName{} will enable structured progress towards fast, efficient CUAs, unlocking new real-world use cases and applications. 
All code, infrastructure, and analysis are available at \href{https://cuaspeedrun.com}{\nolinkurl{cuaspeedrun.com}}.
\end{abstract}

\section{Introduction}

Computer use agents (CUAs) are language model agents that interact with graphical user interfaces (GUIs) to achieve a user-specified goal. CUAs operate over the same user interface as human users, enabling them to potentially automate a much wider variety of tasks without software-specific APIs.

When measured in terms of success alone, recent model releases have pushed the capabilities of CUAs past human baselines. 
The best frontier models today achieve success metrics above human performance on popular CUA benchmarks: Qwen3.8-Max, Claude Mythos 5, GPT-5.5, and Gemini 3.6 Flash report successes ranging from 78.7\% to 86.1\% on OSWorld-Verified, far above the 72.4\% human reference score~\citep{xie2024osworldbenchmarkingmultimodalagents}.
Across long-horizon benchmarks~\citep{aggarwal2026gymanythingturnsoftwareagent,yuan2026osworld20benchmarkingcomputer,jang2026odysseysbenchmarkingwebagents}, frontier models have also demonstrated the ability to achieve strong performance on complex, long-horizon tasks.%

\begin{figure}[t]
  \centering
  \includegraphics[width=\linewidth]{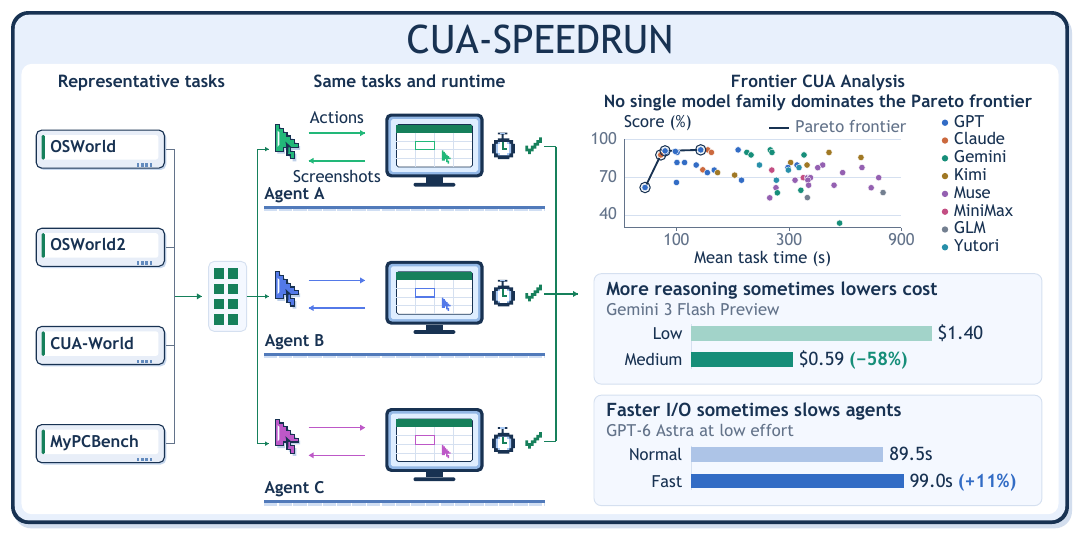}
  \caption{\textbf{\BenchmarkName{} overview.}
\BenchmarkName{} is a standardized platform for benchmarking the performance, speed, and cost of computer-use agents.
It combines representative task selection with standardized infrastructure on Modal, enabling consistent comparisons across models, agent harnesses, and reasoning settings. Our findings reveal that more reasoning can reduce both task time and model cost for CUAs, while faster environment I/O can make agents slower.}
  \label{fig:cua-speedrun-overview}
\end{figure}

Despite their strong performance, a major bottleneck for widespread adoption of current CUAs in practical settings is their speed and cost of deployment~\citep{abhyankar2026osworldhumanbenchmarkingefficiencycomputeruse}.
In particular, we believe that research on the speed and cost efficiency of CUAs has trailed capabilities research, in part due to the difficulty of fairly benchmarking speed and cost of different models in a consistent manner. 
Specifically, CUAs require complex infrastructure to run and evaluate. 
Different benchmarks use varying execution environments, ranging from local virtual machines to cloud-parallelized desktops~\citep{xie2024osworldbenchmarkingmultimodalagents,bonatti2024windowsagentarenaevaluating}. 
There are drastic differences in deployment setup, API interfaces, hyperparameter settings, agent harnesses, and even benchmark implementations across the community.

Towards addressing the problem of consistent benchmarking of CUA wall-time, and igniting research on improving the speed and deployment practicality of CUAs, we propose \BenchmarkName{}: a standardized platform that can use existing benchmarks to measure the speed and cost efficiency of CUAs (Fig.~\ref{fig:cua-speedrun-overview}).
There are several challenges that \BenchmarkName{} addresses, towards standardizing the evaluation of CUAs across different model families (in particular on wall-time metrics):
\begin{itemize}
    \item \textbf{Uniform CUA evaluation infrastructure.} We standardize evaluations through a serverless cloud-based provider\footnote{\url{https://modal.com/}}, where all evaluations run on the exact same execution pipeline and the same virtual machine setup.
    \item \textbf{Common agent interfaces.} Existing CUA implementations are varied and do not always follow the optimal scaffolds. We implement a common agent interface for executing actions and interacting with the environment while allowing for portable agent harnesses and implementations. This allows a single agent implementation to work across different benchmarks. Our analysis also measures time differences due to factors such as the latency of closed-source models, which can vary by provider.
    \item \textbf{A maximally representative subset for fast and comparable evaluations.} CUA benchmarks take a long time to run. We propose an energy minimization framework that selects a minimal representative subset of tasks within existing CUA benchmarks. This allows us to minimize end-to-end evaluation time while preserving the signal of the results, and maintain the relative ordering of agent performance. Our approach reduces evaluation time on OSWorld by 84.9\% while maintaining high correlation with evals run on the full set.
\end{itemize}

We implement these features on \BenchmarkName{}, and benchmark current frontier CUA models on several widely adopted CUA benchmarks, such as OSWorld-Verified~\citep{xie2024osworldbenchmarkingmultimodalagents} and OSWorld 2.0~\citep{yuan2026osworld20benchmarkingcomputer}.
In addition, we also perform evaluations and analysis on CUA-World~\citep{aggarwal2026gymanythingturnsoftwareagent} and MyPCBench~\citep{jang2026mypcbenchbenchmarkpersonallyintelligent}, which test for long-horizon and personalized computer use, respectively.
\BenchmarkName{} allows us to maintain a live Pareto ranking of the most cost-effective, fastest, and most capable models. 
Through an extensive analysis of multiple \BenchmarkName{} runs, we identify several interesting and applicable insights. Success alone hides large differences in speed: on OSWorld, the open-weight Kimi K3 (max reasoning) matches GPT-6 Astra (high), but takes $4.4\times$ longer per task, and none of the open-weight models we evaluate (Kimi K3, MiniMax M3, GLM-5V Turbo) lie on the performance--time or performance--cost Pareto frontier of any benchmarks.
Speed also interacts with reasoning in unintuitive ways: raising reasoning effort can improve benchmark scores while \emph{reducing} task completion time, since the added reasoning avoids repeated unsuccessful actions. Using a fast I/O system can also slow agents, as the models have not been trained to interact with desktops in different I/O speed configurations.

Our findings highlight the importance of turning our attention beyond performance and success rates. \BenchmarkName{} maintains a standardized evaluation of the performance and efficiency of CUAs, two dimensions that we believe are essential future research directions. We hope that the public release of \BenchmarkName{} encourages the research community to work on making CUAs faster, cheaper, and more efficient, paving the way towards more widespread real-world adoption.

\section{Method}

In this section, we describe our approach to address the aforementioned challenges with benchmarking CUAs. We base our techniques to improve CUA benchmarking on standardizing measurements and infrastructure (Sec.~\ref{sec:standardized-cua-eval}), selecting representative tasks instead of running entire benchmarks (Sec.~\ref{sec:representative-tasks}), and building infrastructure that executes and times every agent consistently (Sec.~\ref{sec:infrastructure}).

\subsection{Standardized Computer-Use Evaluation} \label{sec:standardized-cua-eval}

Standardizing computer-use evaluations requires invariance across four different dimensions: (1) the agent, driven by a large-language-model; (2) the benchmark,
which determines what an agent must accomplish; (3) the environment infrastructure, which is responsible for managing VMs, the environment lifecycle, and action/observation contracts; and (4) the agent loop, which determines
how the model interacts with the infrastructure. Many computer-use evaluations often make different choices for these four components, making it difficult to study them independently. For example, adopting a new model may introduce a benchmark-specific
action loop, or adopting a new benchmark requires changes to the agent or infrastructure.
The goal of our setup is to ensure the decoupling of benchmarks from infrastructure from agent design, in order to ensure fair and independent evaluation across these four dimensions.

\paragraph{Problem setup.}
Following prior computer-use benchmarks, each task specifies an initial desktop state, a natural-language instruction, and a verifier, either a programmatic check or an LLM judge, that scores the agent's trajectory and final state. Agents observe screenshots of the desktop and act through keyboard and mouse actions; Appendix~\ref{app:task-definition} gives the formal definition. Comparisons between agents hold the benchmark and infrastructure fixed, and similarly comparisons across benchmarks fix agent and infrastructure, so each change behaves as a clean ablation.

\paragraph{Measuring speed.}
For each task, our infrastructure starts measuring time when the instruction is given to the agent and
stops when the agent terminates or reaches the task limit (steps taken, wall-time, or both).  Environment
provisioning, task setup, agent initialization, and verification occur outside
of this interval. The infrastructure records the total task time and additionally separates it into the time spent executing environment operations and the time spent executing agent operations. Thus, the agent time can be fully attributed to the ``speed'' of the agent, regardless of any time needed to set up CUA evaluations. Additionally,
we also log the total cost of model calls. In addition, infra failures are retried,
while agent failures end the trajectory, which is then scored.

\subsection{Selecting a Representative Set} \label{sec:representative-tasks}

\paragraph{Choosing a representative evaluation set.} A full computer-use evaluation is often expensive because of long trajectories adding to API/GPU cost, and often requires overhead for managing virtual machines (VMs), adding substantial evaluation cost. For example, GPT-5.4 costs approximately
\$4000 on the CUA-World-Long benchmark~\citep{aggarwal2026gymanythingturnsoftwareagent}. We therefore ask the question: can we reduce the size of the
benchmark needed for eval, while still capturing the success rate of the agents we evaluate on the full set?
We want to choose a subset that keeps the individual agents' scores similar to their full benchmark scores and
preserves the ordering of agents' performance.

\paragraph{Selection criteria based on success.}
Our goal is to automatically select a subset $K$ of the benchmark that is representative of the full benchmark. To ensure that the selected subset generalizes beyond the agents used to construct it, we evaluate the selection procedure using leave-one-agent-out validation. For each agent $m$, we construct a subset using the results of only the other agents (M-1) and then compare the held-out agent's score on the selected subset with its score on the full benchmark. \looseness=-1

We found through multiple iterations of leave-one-model-out evaluation that the best method to select tasks was to minimize the energy distance between the distributions of agent
scores on the full benchmark and subset~\citep{Sz_kely_2013}. Specifically, let  $C_{mi}\in[0,1]$ be the partial score of agent $m$ on task $i$, and let
$B_{mi}=\mathbf{1}[C_{mi}=1]$ denote exact completion.  We represent task $i$
by
\begin{equation}
    z_i=(C_{1i},\ldots,C_{Mi},B_{1i},\ldots,B_{Mi})\in\mathbb{R}^{2M}.
\end{equation}

Thus, two tasks are considered similar when agents exhibit similar patterns of partial and exact completion on them. Let $D_{ij}=\lVert z_i-z_j\rVert_2$ and let
$\overline{D}$ be the mean distance between distinct tasks.  For a subset
$S_K$ of $K$ tasks, we minimize
\begin{equation}
\begin{split}
    \mathcal{E}(S_K)=\frac{1}{\overline{D}}\bigg(
       &\frac{2}{KN}\sum_{i\in S_K}\sum_{j=1}^{N}D_{ij}
       -\frac{1}{K^2}\sum_{i,i'\in S_K}D_{ii'} -\frac{1}{N^2}\sum_{j=1}^{N}\sum_{j'=1}^{N}D_{jj'}
    \bigg).
\end{split}
\end{equation}
This objective favors subsets whose tasks represent the performance patterns present in the full benchmark while avoiding redundant tasks with nearly identical patterns. The final term depends only on the full benchmark and is therefore constant during subset selection. We approximately
minimize this objective by starting with 100 task subsets and replacing one
selected task with an unselected task whenever this reduces the objective.
 Thus, we retain the subset with the lowest objective value across the 100 runs. During leave-one-agent-out evaluation, this procedure is repeated after removing the held-out agent’s results. After determining the desired subset size, we construct the final deployed subset once using all available agents.

\paragraph{What subset is sufficient to preserve model rankings?}
The previous selection criteria can find the representative set for a given value of $K$. Next, we ask for the smallest subset of $K$ tasks that consistently preserves
the ranking of agents by performance.  Let $\rho_q(K)$
be the Spearman correlation between leave-one-agent-out estimates and full benchmark
results for evaluation quantity $q$.  We choose
\begin{equation}
    K^*=\min\left\{K:\min_{q}\min_{k\in\{K-1,K,K+1\}}
         \rho_q(k)\geq 0.95\right\}.
\end{equation}
We require the correlation threshold to hold for (K-1), (K), and (K+1) to prevent us from selecting a value of (K) that performs well only by chance.
This criterion selects 50 of 295 OSWorld tasks, 52 of 63 OSWorld2 tasks,
26 of 143 CUA-World tasks, and 38 of 184 MyPCBench tasks
(Table~\ref{tab:representative-task-counts}), which form the
\BenchmarkName{} evaluation sets. We also evaluate how well the subsets preserve full-benchmark performance and model comparisons in our experiments (Appendix~\ref{app:selection}).

\subsection{Infrastructure} \label{sec:infrastructure}

\paragraph{How do we host our infrastructure?} We adapt the virtual machine runtime from
Gym-Anything~\citep{aggarwal2026gymanythingturnsoftwareagent} to run task environments natively
in Modal sandboxes. Our infrastructure manages separate sandboxes for agents
and environments, ensuring isolation, while Modal sandboxes provide a
standardized hosted runtime so that differences in users' infrastructure do
not affect evaluation results. For self-hosted open-weight models, we use vLLM~\citep{kwon2023efficientmemorymanagementlarge} inference
servers on fixed L40S GPUs. Keeping the inference servers and GPUs fixed allows consistency in evaluation. We develop a Python library to handle GPU scheduling,
environment allocation, and parallelization, with further details in
Appendix~\ref{app:infrastructure}.

\paragraph{Action and observation modalities.} Following standard practice in computer-use agents~\citep{xie2024osworldbenchmarkingmultimodalagents,qin2025uitarspioneeringautomatedgui}, the action space consists of keyboard actions (e.g., typing text or pressing
\texttt{Ctrl+C}) and mouse actions (e.g., clicking at coordinates $(x,y)$ or
double-clicking). Agent observations are RGB screenshots of the desktop at a
resolution of $1920\times1080$. 

To ensure that these interfaces behave consistently across VM runtimes, we
developed \textit{CUA-AutoDebug}, an end-to-end test suite in which a
controlled application records the keyboard and mouse inputs it receives,
compares them with the intended inputs for a large catalog of actions, and
checks that screenshots match the application state. These tests uncovered
several input errors in widely used CUA tools, which we correct in our
infrastructure (Appendix~\ref{app:autoharness}).

\paragraph{FastCUA: Developing a new fast I/O system for CUAs.} Standard computer-use runtimes
typically take roughly 2--3 seconds from issuing an action to receiving an
observation. This latency is dominated by a variety of elements, such as action execution, waiting for the
application to respond, and networking. The delay incidentally gives the application
time to process the action (e.g., open a menu), allowing the agent to observe
its effect. We therefore follow this design for our main evaluations.
However, we develop a fast I/O mode, \textit{FastCUA}, by optimizing networking, action
execution, and image processing latencies (see Appendix~\ref{app:infra-latency}). FastCUA reduces action-to-observation latency to 2--28\,ms, an improvement of more than an order of magnitude over typical runtimes. At this speed, a screenshot can be captured before the application has updated its UI in response to an action. We evaluate this setting to test whether lower infrastructure latency reduces overall wall time.

\paragraph{Implementing agents on \BenchmarkName{}.} Since we have standardized infrastructure, new agents can be implemented with a single Python file that contains the agent loop. We also provide standard templates for various popular agents, which users can tweak directly. We provide options for both hosted evaluations and local evaluations through a single CLI call. See Appendix~\ref{app:agent-guide} for more details.

\section{Results}
\label{sec:results}

\subsection{Experimental Setup} \label{sec:experimental-setup}

We evaluate 56 agent configurations on OSWorld and 21 on OSWorld2 using the
representative task sets from Sec.~\ref{sec:representative-tasks}. These
configurations cover frontier open-weight and proprietary models, each run
through its public reference agent implementation or native computer-use API
at selected reasoning-effort settings, with the tasks and infrastructure
held fixed (Appendix~\ref{app:experimental-specification}). We validate all agent designs for correctness
through manual inspection and CUA-AutoDebug on representative tasks. For selected models, we also compare different agent
harness designs, standard and fast I/O, and single versus batched tool calls
per model response. Repeated evaluations are stable: across five seeds,
GPT-6 Astra (low) scores $90.8 \pm 1.8$\% on OSWorld with a mean task time
of $89.5 \pm 2.2$\,s (Appendix~\ref{app:run-variation}).

\paragraph{Metrics.}
For each configuration, we report the mean verifier score, which retains
partial credit, together with the mean task time
(Sec.~\ref{sec:standardized-cua-eval}) and mean model cost per task. We report
the performance--time and performance--cost Pareto frontiers, as well as the
joint frontier over all three. Interaction turns, generated tokens, and exact completion
are defined in Appendix~\ref{app:measurements}.

\begin{figure}[!t]
  \centering
  \includegraphics[width=\linewidth]{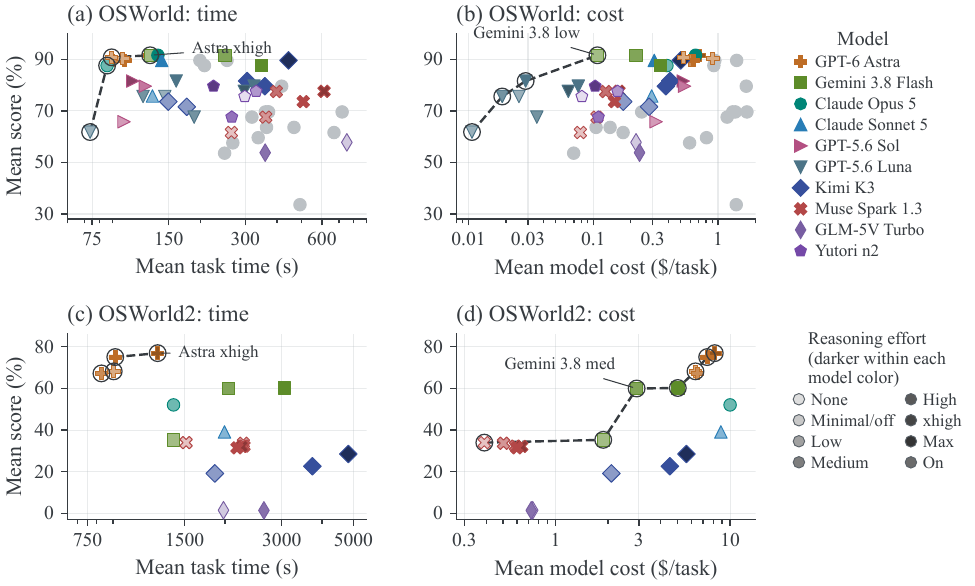}
  \caption{\textbf{Different agents define the speed and cost frontiers.}
  Mean score vs. mean task time and model cost on OSWorld (top, 54
  configs) and OSWorld2 (bottom, 19 configs), excluding MiniMax M3. Dashed lines
  connect the observed Pareto-frontier points, which are circled. Colors and marker shapes
  identify models, and darker shades indicate increased reasoning effort.
  The plots compare complete agent configs. Costs use recorded charges or
  token usage at the applicable API prices. Astra (low) uses five-seed means for each I/O setting.
  Full configs and plots in Appendix~\ref{app:full-results}.}
  \label{fig:results-overview}
\end{figure}

\subsection{Analysis}

\paragraph{No single model family dominates the frontier.}
Fig.~\ref{fig:results-overview} shows the performance--time and
performance--cost Pareto frontiers on both benchmarks. On OSWorld, the
high-performance end of the time frontier extends from Claude Opus 5 (low),
with a score of 87.6\% in 86\,s, through GPT-6 Astra (low), with 90.8\% in
90\,s, to Astra (xhigh), with 91.6\% in 127\,s. Gemini 3.8 Flash (low) matches the highest score and takes almost
the same time as GPT-6 Astra, but costs \$0.11 rather than \$0.71 per task.
At the lower-cost end, GPT-5.6 Luna (low, direct API) costs about \$0.01
per task but scores 61.8\%. Thus, \BenchmarkName{} enables
comparison across various dimensions such as speed, cost and performance, rather than
reducing the comparison to a single model ranking.

\paragraph{The Pareto frontier changes across benchmarks.}
Fig.~\ref{fig:results-overview} also shows that, unlike the OSWorld time frontier, the OSWorld2 time frontier
is formed entirely by GPT-6 Astra configurations. Astra (high) achieves
75.0\% in 914\,s, while xhigh reaches 76.9\% in 1,237\,s, achieving an
additional 1.9 percentage points with 35\% more time. However, the cost frontier includes
other models such as Gemini 3.8 Flash (medium), which achieves 59.9\% at
\$2.91 per task, compared to Astra (high) at \$7.39. Muse Spark 1.3
(minimal) extends the frontier to lower cost, with a score of 34.0\%
at an estimated \$0.39. Surprisingly, there is no open model on either the cost or the time frontier for either OSWorld or OSWorld2. Tracking these frontiers allows us to highlight the trade-off
explicitly. The preferred agent depends on both the benchmark and the
performance required within a time or cost budget, and there is no universal model family that dominates across different benchmarks.
\begin{figure}[!t]
  \centering
  \includegraphics[width=\linewidth]{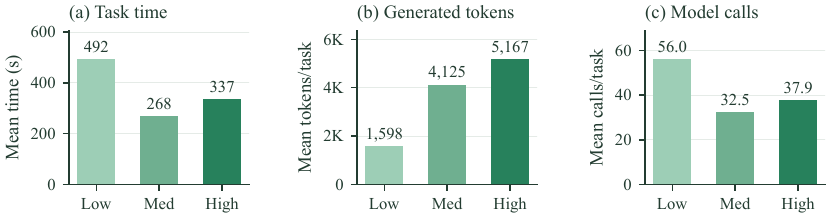}
  \caption{\textbf{Less reasoning can make an agent slower.}
  Gemini 3 Flash Preview at low, medium, and high effort on OSWorld. The panels show mean task time,
  generated tokens including reasoning, and model calls across all 50 tasks.
  Scores are 33.6\%, 57.6\%, and 59.6\%, respectively. Medium and high effort
  generate more tokens than low effort but require fewer calls and
  complete tasks faster.}
  \label{fig:results-reasoning-interactions}
\end{figure}

\begin{figure}[!t]
  \centering
  \includegraphics[width=\linewidth]{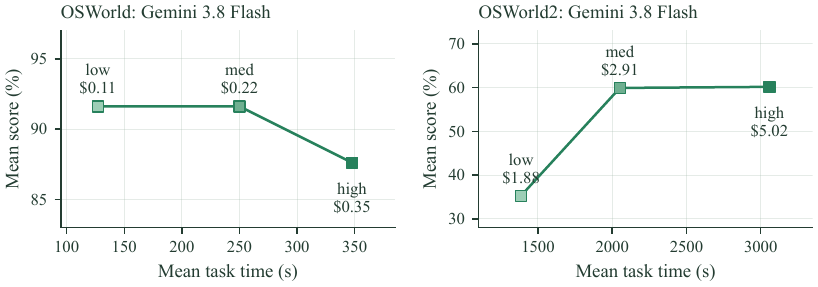}
  \caption{\textbf{The useful reasoning setting changes across benchmarks.}
  Gemini 3.8 Flash at low, medium, and high effort; labels also report mean
  cost per task. Each curve keeps the model, agent harness, and execution
  settings fixed. Additional reasoning brings no gain on OSWorld, whereas
  moving from low to medium substantially improves OSWorld2 performance.}
  \label{fig:results-reasoning}
\end{figure}

\begin{figure}[!t]
  \centering
  \includegraphics[width=\linewidth]{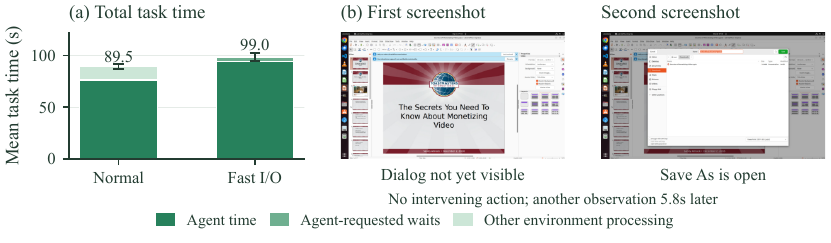}
  \caption{\textbf{Faster I/O can be offset by more waiting and agent interaction.}
  GPT-6 Astra (low), with normal and fast I/O. Panel (a) separates agent
  time, agent-requested waits, and remaining environment processing,
  averaged over five seeds per setting; error bars show the standard
  deviation of mean task time across seeds. Panel (b) shows two consecutive
  screenshots after opening Save As: the dialog is absent in the first
  and visible in the second, requested 5.8\,s later with no intervening action.}
  \label{fig:results-fast-io}
\end{figure}

\paragraph{Less reasoning can sometimes make agents slower.}
Fig.~\ref{fig:results-reasoning-interactions} compares Gemini 3 Flash Preview
at different reasoning settings on OSWorld, keeping the model and harness
fixed. Moving from low to medium effort improves the score from 33.6\% to
57.6\% while reducing mean task time from 492\,s to 268\,s. We diagnose
this as follows: at lower reasoning effort, while the model generates fewer tokens (1,598 vs.\  4,125 at medium effort), it takes substantially more steps (56.0 vs.\ 32.5),
since most of its steps are suboptimal. For instance, the trajectories show repeated unsuccessful actions
at low effort, such as repeatedly trying to enter spreadsheet text through
key combinations instead of text entry. Counterintuitively, higher reasoning effort can therefore
shorten the interaction enough to reduce total task time, even when it
increases the number of generated tokens per step.

\paragraph{More reasoning can increase time and cost without improving performance.}
Fig.~\ref{fig:results-reasoning} compares reasoning settings within
Gemini 3.8 Flash on each benchmark. On OSWorld, low and medium reasoning both score
91.6\%, but medium increases mean task time from 127\,s to 250\,s and
approximately doubles cost. However, the trend changes on OSWorld2: moving from low to medium
improves performance from 35.3\% to 59.9\%. But increasing effort further
to high adds only 0.2 percentage points while increasing time by 49\%
and cost by 72\%. GPT-6 Astra shows a similar dependence on the benchmark: on OSWorld, medium effort is within
2 points of xhigh at 29\% less time and 15\% less cost, whereas on OSWorld2,
high effort gains 7.8 points over medium for 10\% more time.
The useful reasoning setting therefore depends on the
tasks as well as the model: additional reasoning can be valuable on one
(harder) benchmark and unnecessary on another (simpler) benchmark.

\paragraph{Faster infra can make the agent slower.}
Fig.~\ref{fig:results-fast-io} compares GPT-6 Astra (low) with the normal input-output mode and
our faster input-output mode, FastCUA (Sec.~\ref{sec:infrastructure} and Appendix~\ref{app:fast-io-profile}) across five seeds per setting. Interestingly, while the
mean environment processing time falls from 12.68\,s to 0.34\,s per task,
total task time increases from 89.5\,s to 99.0\,s. We also find
that the number of steps per trajectory increases in FastCUA. This is because fast I/O can return screenshots before an application
has updated its interface. For example, after opening Save As, the agent
receives a screenshot without the dialog and requests another screenshot
5.8\,s later, which shows it open. Although theoretically a model could account for such faster I/O by using appropriate wait times
(e.g., 100 \, ms in this case) for
the application to update, we find that current models are not able to optimally utilize this, resulting in much higher times.

\paragraph{The agent harness plays an important role in the speed--performance trade-off.}
Fig.~\ref{fig:results-harness-full} compares GPT-5.6 Luna through the direct-API agent and the Codex harness on
OSWorld.
At low effort, Codex improves the score from 61.8\% to 75.6\%, but increases mean task time from 74\,s to 145\,s. The same model and reasoning setting therefore produce different performance and efficiency when used through different harnesses. At medium and high effort, the direct-API agent is instead both more accurate and faster (75.6\% in 119\,s vs.\ 67.6\% in 189\,s at medium; 81.6\% in 161\,s vs.\ 79.6\% in 325\,s at high), so neither harness is uniformly better.

\paragraph{Batching actions reduces task time.}
Each step costs a screenshot, a model call, and environment latency, so
agents that issue several actions per response (which we execute
sequentially) finish sooner: GPT-6 Astra (low) completes OSWorld tasks in
5.9 action batches on average (Table~\ref{tab:run-variation}). Restricting
GPT-6 Astra (xhigh) to one action per response keeps its score at 91.6\%
but raises mean task time from 127\,s to 183\,s. The benefit depends on
how well a model plans multi-action sequences, and batching alone does not
make an agent fast (Appendix~\ref{app:batching}).

\paragraph{Fast agents generate fewer tokens, not tokens faster.}
Claude Opus 5 (low) generates 1,795 tokens per task compared with 1,579 for
GPT-6 Astra (low), while scoring 87.6\% vs.\ 90.8\%. At a common token rate
(Fig.~\ref{fig:full-normalized}), it falls behind the Astra configurations,
which remain the fastest high-scoring agents. Similarly, Claude Sonnet 5
(high) takes 141\,s for the score that Astra (medium) reaches in 90\,s,
generating $3.5\times$ more tokens, and Gemini 3.8 Flash (high) generates
$5.6\times$ more tokens than at low effort and finishes $2.7\times$ later.

\paragraph{Extending to other benchmarks.}
We additionally evaluate GPT-6 Astra with Codex at four reasoning settings on
MyPCBench~\citep{jang2026mypcbenchbenchmarkpersonallyintelligent} and
CUA-World~\citep{aggarwal2026gymanythingturnsoftwareagent}, which test
personalized and long-horizon computer use and grade trajectories with
vision-language models rather than programmatic checks. Astra exceeds 90\%
rubric scores on both (Appendix~\ref{app:additional-benchmarks}), highlighting
the need for more challenging task sets.

\subsection{Practical Suggestions for Building CUAs}
\begin{figure}[!t]
  \centering
  \includegraphics[width=\linewidth,trim=0 0 0 24bp,clip]{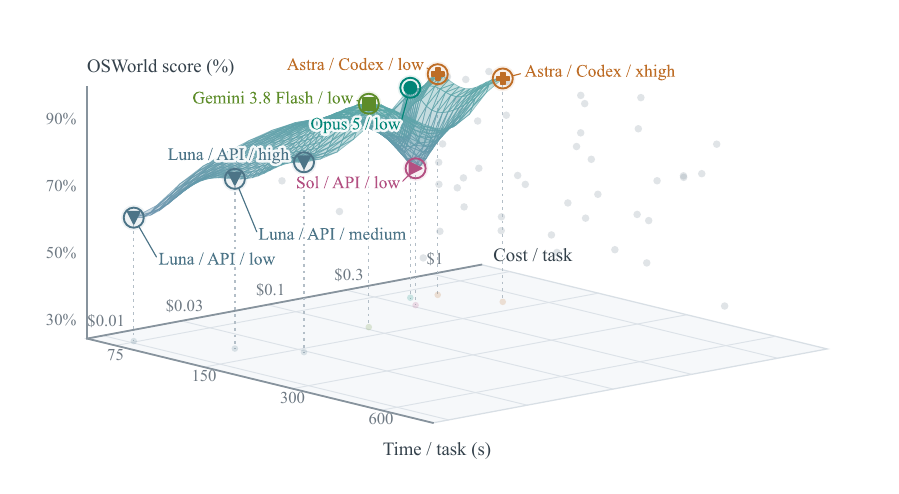}
  \vspace{-6ex}
  \caption{\textbf{Choosing an agent by performance, time, and cost.}
  Joint Pareto frontier on OSWorld; the OSWorld2 frontier is in
  Appendix~\ref{app:full-results}, Fig.~\ref{fig:joint-osworld2}. Circled points are the
  configurations on the Pareto frontier. Time and cost
  axes are logarithmic. \looseness=-1}
  \label{fig:results-joint-frontier}
\end{figure}

Based on the analysis above, we make the following suggestions for building
efficient CUAs:
\begin{itemize}
    \item \textbf{Choose a frontier configuration for the required performance
    and budget.} No single configuration is best on every metric, and the
    frontier shifts across benchmarks (Figs.~\ref{fig:results-overview}
    and~\ref{fig:results-joint-frontier}), so candidates should be measured on
    tasks that resemble the target workload.
    \item \textbf{Tune reasoning effort rather than setting it to an extreme.}
    Too little reasoning can lengthen trajectories, while too much can add time
    and cost without improving performance
    (Figs.~\ref{fig:results-reasoning-interactions}
    and~\ref{fig:results-reasoning}). Start with low or medium effort and raise
    it only for a measured gain.
    \item \textbf{Batch actions when intermediate screenshots are unnecessary.}
    Fewer steps mean fewer screenshots, model calls, and environment round
    trips.
    \item \textbf{Reduce generated tokens and steps rather than maximizing
    token rate or I/O speed.} Faster generation or I/O helps only if the agent
    does not spend it on longer outputs or additional steps
    (Fig.~\ref{fig:results-fast-io}).
    \item \textbf{Optimize the harness alongside the model.} The same model and
    reasoning setting can land at different points on the frontier depending
    on its harness (Fig.~\ref{fig:results-harness-full}).
\end{itemize}

\section{Related Work}

\paragraph{Computer-use agent benchmarks.}
The first computer-use agent benchmarks used synthetic interfaces~\citep{liu2018reinforcementlearningweb,yao2022webshop}. Follow-up evaluations moved to self-hosted realistic website analogs~\citep{zhou2024webarenarealisticwebenvironment,koh2024visualwebarenaevaluatingmultimodalagents,drouin2024workarena}, static datasets of real live website tasks~\citep{deng2023mind2web}, and live-Internet evaluations~\citep{he2024webvoyager,xue2025illusionprogressassessingcurrent}.
Live, dynamic desktop benchmarks such as OSWorld~\citep{xie2024osworldbenchmarkingmultimodalagents} evaluate agents on real Linux desktops using programmatic verifiers, with WindowsAgentArena~\citep{bonatti2024windowsagentarenaevaluating}, macOSWorld~\citep{yang2025macosworld}, iOSWorld~\citep{jang2026iosworldbenchmarkpersonallyintelligent}, and AndroidWorld~\citep{rawles2025androidworld} extending coverage to other platforms and professional workflows.
Recent benchmarks stress long horizons and scale, such as OSWorld 2.0~\citep{yuan2026osworld20benchmarkingcomputer}, CUA-World from Gym-Anything~\citep{aggarwal2026gymanythingturnsoftwareagent}, and Odysseys~\citep{jang2026odysseysbenchmarkingwebagents}. Another emphasis has become personal assistant evaluations, including OpenClaw-style works such as ClawBench~\citep{zhang2026clawbench}, WildClawBench~\citep{ding2026wildclawbench}, and MyPCBench~\citep{jang2026mypcbenchbenchmarkpersonallyintelligent}, which evaluate personalized computer use and digital assistants. OSWorld-Human~\citep{abhyankar2026osworldhumanbenchmarkingefficiencycomputeruse} measures agent latency on OSWorld and compares step counts against human reference trajectories, finding that agents take longer than humans.

\paragraph{Computer-use agent models.}
Frontier model releases have increasingly prioritized computer use~\citep{openai2026gpt6astra,anthropic2026fablemythos,google2026gemini36flash,qwen38}, with several reporting above-human scores on OSWorld-Verified. Models such as Kimi K3~\citep{kimi2026k3}, GLM-5~\citep{glm2026glm5}, and Muse Spark~\citep{meta2026musespark} have also included agentic workflows in their tech releases. Alongside these large general models, a line of work trains specialized computer-use models, including UI-TARS~\citep{qin2025uitarspioneeringautomatedgui}, MolmoWeb~\citep{gupta2026molmowebopenvisualweb}, Fara-1.5~\citep{awadallah2026fara15}, ScaleCUA~\citep{lv2026scalecua}, and Qwen-CUA~\citep{lu2026qwencua}, often in verifiable environments generated at scale~\citep{wang2026cuagymscalingverifiabletraining,aggarwal2026gymanythingturnsoftwareagent}.

\paragraph{Reproducible and efficient evaluation.}
Agent evaluations are noisy~\citep{kapoor2024aiagentsmatter,hal}, as \citet{xue2025illusionprogressassessingcurrent} showed that reported web-agent success is inflated by evaluation issues, and \citet{sahu2026teachittostop} show that single-run claims are unreliable because outcomes vary substantially across data draws and run non-determinism. A separate line of work reduces the evaluation cost by selecting a small subset of a benchmark that predicts full-benchmark scores, via item response theory~\citep{polo2024tinybenchmarks,kipnis2025metabench,zhou2026lostinbenchmarks}, sparse optimization~\citep{zhang2026sparseeval}, or human validation~\citep[SWE-bench Verified;][]{chowdhury2024swebenchverified}. PACE~\citep{song2026pace} predicts agentic benchmark scores from cheap, non-agentic evaluations.

\section{Conclusion}

In this work, we introduced \BenchmarkName{}, a framework for evaluating
the performance, speed, and cost of computer-use agents under standardized
infrastructure. Our evaluations on OSWorld and OSWorld2 identify the
performance--time and performance--cost Pareto frontiers and show how these
change across benchmarks. We further find that more reasoning can reduce
task time by avoiding repeated unsuccessful actions, while faster I/O can
increase it through additional interaction. These results highlight the
need to study reasoning and interaction together when developing faster
agents. We hope \BenchmarkName{} enables the community to improve both
capability and efficiency, making computer-use agents more practical
to deploy.

\subsubsection*{Acknowledgements}
We thank Modal for their generous support in cloud compute credits to build \BenchmarkName{}. We thank Eunsu Kim, Naveen Raman, Mareks Woodside, Seungone Kim, Zeyu Zheng, and others for feedback and helpful discussions. Jing Yu Koh is supported by a Jane Street Graduate Research Fellowship. Lawrence Jang is supported by a Susquehanna International Group PhD Fellowship. Pranjal Aggarwal is supported by a SoftBank Group-Arm Fellowship. This work is partially supported by the National Science Foundation under Grant No. DMS-2502281, and a grant from Amazon on Useful Chain of Thought Reasoning. 

\subsubsection*{AI use statement}
We used AI assistance for general software coding, making figures, polishing writing across the manuscript, and verifying related work.

\bibliography{references}
\bibliographystyle{iclr2027_conference}

\appendix
\clearpage
\raggedbottom
\hypersetup{hidelinks}
\startcontents[appendices]
\begingroup
\hypersetup{colorlinks=true,linkcolor=appendixgreen}
\titlecontents{section}[1.6em]
  {\Needspace{6\baselineskip}\addvspace{0.65em}\bfseries}
  {\contentslabel{1.6em}}{}
  {\titlerule*[0.6pc]{.}\contentspage}
\titlecontents{subsection}[4.0em]
  {\small}
  {\contentslabel{2.4em}}{}
  {\titlerule*[0.6pc]{.}\contentspage}
\section*{Appendix Table of Contents}
\printcontents[appendices]{}{1}{\setcounter{tocdepth}{2}}
\endgroup
\clearpage

\section{Limitations}
\label{app:limitations}

Reported times and costs reflect API inference conditions at the time of
evaluation (Appendix~\ref{app:experimental-specification}). Providers control
subsequent changes to inference speed and pricing. We validate representative
task subsets on held-out agents (Appendix~\ref{app:selection}); subset scores
approximate full-benchmark performance. Changes in agent capabilities may
require revalidation of these subsets.

Our evaluations cover a broad range of agents across four benchmarks
(Appendix~\ref{app:full-results}). Evaluation cost limits coverage of every
combination of model, harness, reasoning effort, and benchmark.

\section{Infrastructure: Technical Details}
\label{app:infrastructure}
\label{app:infra-latency}

\subsection{Agent and Environment Sandboxes}

We run the agent and task environment in separate sandboxes. The
environment contains the desktop, applications, and initial task state;
the agent interacts with it through screenshot and action requests.
Task initialization and verification remain independent of the agent,
allowing the same agent implementation to run across benchmarks.
When the agent terminates or reaches its limit, the benchmark's verifier
scores the trajectory or final environment state.

We use Modal to host the sandboxes and adapt the environment runtime
from Gym-Anything~\citep{aggarwal2026gymanythingturnsoftwareagent}.
The environment measures task time and the time spent executing actions
and returning observations, independently of the agent implementation.
As in Sec.~\ref{sec:standardized-cua-eval}, task preparation, agent
initialization, and verification are excluded from task time.

\subsection{Model Serving, Resource Allocation, and Parallel Evaluation}

Agents access models through API endpoints or a self-hosted vLLM
inference server~\citep{kwon2023efficientmemorymanagementlarge}.
We use L40S GPUs for self-hosted inference.
The evaluation configuration specifies the compute resources, task
limits, and number of concurrent agents. Model initialization and
weight loading finish before task timing begins.

We assign tasks to available agents and environments. Each task starts
from a fresh environment state. Its clock starts after the agent and
environment are ready and stops when the agent finishes or reaches its
limit, excluding time waiting for an execution slot. Model servers reuse
loaded weights across tasks. We record a trajectory for each task.

\subsection{Fast I/O Implementation}
\label{app:fast-io-profile}

FastCUA, our Fast I/O implementation, reduces the time between issuing
an action and receiving a screenshot. We encode screenshots in the
background and combine action execution and screenshot delivery in a
single request.

\paragraph{Screenshot capture and encoding.}
We use QEMU's D-Bus display interface to keep the current screen in memory
and encode updated frames in the background.
We use JPEG at quality 95 with no chroma subsampling and retain the
full $1920\times1080$ resolution.

\paragraph{Action execution and communication.}
We execute the action and return its screenshot in one request, reusing
the network connection across requests. A lightweight command client
reduces per-action startup overhead. We also place
agent and environment sandboxes in the same Modal region to reduce
network latency. Fast I/O returns the screenshot as soon as action
execution and image preparation finish.
It can therefore return a screenshot before the application has updated
its interface, as illustrated in Figure~\ref{fig:results-fast-io}.

\FloatBarrier
\subsection{Action Latency: Fast I/O, Gym-Anything, and OSWorld}

\begin{table}[H]
  \centering
  \small
  \begin{tabular}{lrrrr}
    \toprule
    Infrastructure & Click & Escape & Ctrl+U & Type 100 characters \\
    \midrule
    Gym-Anything & 2758.70 & 2865.00 & 2858.29 & 3521.72 \\
    OSWorld & 2515.41 & 2707.74 & 2708.41 & 2756.47 \\
    Fast I/O (ours) & \textbf{27.58} & \textbf{2.43} & \textbf{2.91} & \textbf{14.98} \\
    \bottomrule
  \end{tabular}
  \caption{\textbf{Fast I/O reduces action-to-observation latency to milliseconds.}
  Median latency in milliseconds over 20 calls per action. For each system,
  the timer starts immediately before \texttt{env.step(action)} and stops
  when it returns the observation. All default action waits are included.}
  \label{tab:infra-latency}
\end{table}

\Needspace{10\baselineskip}
\paragraph{Fast I/O reduces latency by one to three orders of magnitude.}
Table~\ref{tab:infra-latency} compares Fast I/O with
Gym-Anything~\citep{aggarwal2026gymanythingturnsoftwareagent} and
OSWorld~\citep{xie2024osworldbenchmarkingmultimodalagents} for mouse
clicks, key presses, and text entry.
Fast I/O takes 2.4--27.6\,ms per call, compared to 2.5--3.5\,s for the
existing implementations. For example, a click takes 27.6\,ms compared with
2.8\,s with Gym-Anything, while typing 100 characters takes 15.0\,ms
compared with 3.5\,s.

\paragraph{Measurement setup.}
We time each runtime's native \texttt{env.step(action)} call, from issuing
the action until the observation is returned. We test four actions on a
$1920\times1080$ Ubuntu desktop: clicking the Activities button, pressing
Escape, pressing Ctrl+U, and typing 100 ASCII characters into a terminal.
The caller and VM run in the same Modal sandbox. OSWorld uses its Docker
provider with screenshot observations; both
baselines use their default waits and action implementations.
Table~\ref{tab:infra-latency} reports the median of 20 calls per action,
without profiling. Setup, pauses between calls, and saving returned
screenshots occur outside the timed interval.

\FloatBarrier
\subsection{Latency Breakdown}

\begin{table}[H]
  \centering
  \small
  \begin{tabular}{lrrrr}
    \toprule
    Component (ms) & Click & Escape & Ctrl+U & Type 100 characters \\
    \midrule
    \multicolumn{5}{l}{\textbf{Gym-Anything}} \\
    Fixed waits & 2100.2 & 2100.2 & 2100.2 & 2100.2 \\
    Action request & 153.5 & 153.7 & 154.0 & 805.5 \\
    Screenshot capture & 258.9 & 354.0 & 354.3 & 354.1 \\
    Screenshot transfer & 238.1 & 212.4 & 220.6 & 211.3 \\
    Cleanup and other & 53.1 & 53.1 & 53.1 & 53.1 \\
    \textbf{Total} & \textbf{2803.9} & \textbf{2873.4} & \textbf{2882.2} & \textbf{3524.2} \\
    Total minus 2\,s & 803.9 & 873.4 & 882.2 & 1524.2 \\
    \midrule
    \multicolumn{5}{l}{\textbf{OSWorld}} \\
    Fixed wait & 2000.1 & 2000.1 & 2000.1 & 2000.1 \\
    Action request & 198.6 & 191.7 & 192.8 & 245.1 \\
    Screenshot request & 324.7 & 521.2 & 521.8 & 523.2 \\
    Other & 0.2 & 0.2 & 0.2 & 0.2 \\
    \textbf{Total} & \textbf{2523.6} & \textbf{2713.3} & \textbf{2714.9} & \textbf{2768.7} \\
    Total minus 2\,s & 523.6 & 713.3 & 714.9 & 768.7 \\
    \bottomrule
  \end{tabular}
  \caption{\textbf{Fixed waits account for most baseline latency.}
  Mean component times from 10 profiled calls per action for Gym-Anything
  and 20 for OSWorld. Components sum to the total before rounding.
  The final row for each system is the profiled total minus its two-second
  post-action wait.}
  \label{tab:baseline-io-profile}
\end{table}

\paragraph{Where do the baselines spend time?}
Table~\ref{tab:baseline-io-profile} shows that fixed sleeps account for
most of the baseline latency. Both implementations sleep for two seconds
after an action, and Gym-Anything adds another 100\,ms per step.
Subtracting the two-second sleep leaves 804--1524\,ms for
Gym-Anything and 524--769\,ms for OSWorld. Gym-Anything captures a PNG
with FFmpeg, transfers it, and deletes the temporary file; typing also
adds a 6\,ms interval between characters. In OSWorld, most of the
remaining time is spent in the action and screenshot requests, including
processing inside the VM and communication.

\begin{table}[H]
  \centering
  \small
  \begin{tabular}{lrrrr}
    \toprule
    Component (ms) & Click & Escape & Ctrl+U & Type 100 characters \\
    \midrule
    Input execution & 27.30 & 1.85 & 2.55 & 18.39 \\
    Frame capture & 1.03 & 0.77 & 0.77 & 0.80 \\
    Foreground image preparation & 0.27 & 0.02 & 0.02 & 0.18 \\
    Other environment processing & 0.93 & 0.94 & 1.00 & 0.90 \\
    Transport & 4.67 & 4.99 & 4.76 & 4.62 \\
    Screenshot-file delivery & 0.43 & 0.45 & 0.43 & 0.44 \\
    Command overhead & 4.41 & 4.52 & 4.38 & 4.38 \\
    \midrule
    \textbf{Total} & \textbf{39.06} & \textbf{13.55} & \textbf{13.90} & \textbf{29.71} \\
    \bottomrule
  \end{tabular}
  \caption{\textbf{Fast I/O completes the four tested actions in under 50\,ms.}
  Mean times over 20 commands per action, from the agent issuing a command
  to receiving the screenshot file. This includes command overhead,
  communication between sandboxes, and screenshot-file delivery in addition
  to the \texttt{env.step} call.
  Components sum to the total before rounding.}
  \label{tab:fast-io-profile}
\end{table}

\FloatBarrier
\paragraph{Full command-to-screenshot latency.}
Table~\ref{tab:fast-io-profile} measures latency from the agent issuing a
command to receiving the screenshot file, including communication
between sandboxes. We place both sandboxes in New York and test the four
actions on two OSWorld desktops, with 0.5\,s and 8\,s gaps between
requests. All 80 timed commands complete in under 50\,ms.
Median times are 38.5\,ms for a click, 12.8\,ms for
Escape, 13.6\,ms for Ctrl+U, and 29.5\,ms for typing 100 characters.
Communication contributes 4.6--5.0\,ms on average. Image preparation
after the request takes 0.02--0.27\,ms; encoding runs in the background.
Clicks take longer because input execution
includes a 25\,ms interval between pressing and releasing the mouse button.

\FloatBarrier
\subsection{Fast I/O Consistency Across Five Seeds}
\label{app:fast-io-consistency}

\begin{table}[H]
  \centering
  \footnotesize
  \setlength{\tabcolsep}{4pt}
  \begin{tabular}{lrrrrrrr}
    \toprule
    Run & Score (\%) & Time (s) & Agent (s) & Steps & Env. (s) & Waits (s) & Env.$-$waits (s) \\
    \midrule
    1 & 91.62 & 103.96 & 98.07 & 8.68 & 5.891 & 5.553 & 0.338 \\
    2 & 91.62 & 93.98 & 89.02 & 8.04 & 4.955 & 4.628 & 0.327 \\
    3 & 91.62 & 96.48 & 91.67 & 7.96 & 4.812 & 4.488 & 0.324 \\
    4 & 87.62 & 100.22 & 95.28 & 8.20 & 4.934 & 4.604 & 0.330 \\
    5 & 89.62 & 100.46 & 95.68 & 8.12 & 4.785 & 4.424 & 0.361 \\
    Mean & 90.42 & 99.02 & 93.94 & 8.20 & 5.075 & 4.739 & 0.336 \\
    SD & 1.79 & 3.87 & 3.58 & 0.28 & 0.462 & 0.462 & 0.015 \\
    \bottomrule
  \end{tabular}
  \caption{\textbf{Fast I/O remains consistent across five evaluations.}
  Each row averages all 50 OSWorld tasks for GPT-6 Astra at low reasoning
  effort. SD is the sample standard deviation across run-level means;
  score SD is in percentage points. Steps count action batches, and
  waits are explicitly requested by the agent.}
  \label{tab:fast-io-variation}
\end{table}

\paragraph{Environment processing time remains low across runs.}
Table~\ref{tab:fast-io-variation} reports five evaluations of GPT-6 Astra
at low effort on the same 50 OSWorld tasks with Fast I/O. We keep the
agent prompt fixed and use fresh task seeds, a 500-step limit, and agent
and environment sandboxes in New York.
Environment time excluding agent-requested waits is
$0.336 \pm 0.015$\,s per task (mean $\pm$ standard deviation across runs).
The mean task time is $99.0 \pm 3.9$\,s, and the score is
$90.4 \pm 1.8$\%.

\paragraph{Most environment time is spent on agent-requested waits.}
Table~\ref{tab:fast-io-variation} also separates the waits requested by the
agent from other environment processing. These waits account for
$4.739$\,s of the $5.075$\,s of mean environment time per task, leaving
only $0.336$\,s for the remaining processing.
Appendix~\ref{app:timing-components} defines these timing components.

\FloatBarrier
\Needspace{8\baselineskip}
\section{Implementing and Evaluating an Agent}
\label{app:agent-guide}

To evaluate a new agent with \BenchmarkName{}, users implement its model
calls, prompts, and interaction history. The infrastructure handles task
setup, desktop interaction, and verification through the common interface
in Sec.~\ref{sec:infrastructure}. The same agent implementation and
environment interface can then be used across benchmarks.

\subsection{Agent Templates}

We provide templates for direct model-API agents, Codex, Claude Code,
and open-weight agents served through vLLM. Each template contains two
files: \texttt{init.py} prepares dependencies and model serving, and
\texttt{agent.py} implements the agent loop.
Initialization finishes before task timing begins. Users can retain
\texttt{init.py} and modify only the agent loop to test a new model,
prompt, or interaction strategy.

\subsection{Minimal Agent Example}

Listing~\ref{lst:agent-loop} shows a minimal agent. The script receives the
environment URL and task instruction as command-line arguments and uses
the \texttt{Computer} client to request screenshots, execute actions,
and end the task. In each iteration, \texttt{choose\_actions} receives
the instruction, screenshot, and interaction history. This user-defined
function calls the model and returns an action list, or \texttt{None}
when the agent considers the task complete. Users determine how to
present this information to the model.

\begin{agentlisting}{Minimal agent loop.\label{lst:agent-loop}}
import os
import sys
from cua_speedrun.client import Computer

def run(env_url, task):
    computer = Computer(env_url)
    history = []
    limit = int(os.environ.get("CS_MAX_STEPS", "500"))
    for _ in range(limit):
        observation = computer.observe()
        actions = choose_actions(task, observation["png"], history)
        if actions is None:
            break
        computer.step(actions)
        history.append((observation["png"], actions))
    computer.done()

if __name__ == "__main__":
    run(sys.argv[1], sys.argv[2])
\end{agentlisting}

For example, an action list can contain
\texttt{\{"mouse": \{"left\_click": [640, 400]\}\}} followed by
\texttt{\{"keyboard": \{"text": "hello"\}\}}.
These two actions are executed in order within one step.
Convenience methods such as \texttt{click(640, 400)},
\texttt{type\_text("hello")}, and \texttt{keys(["ctrl", "c"])}
each submit a single action. The agent can execute several actions before
requesting another screenshot. Calling \texttt{done()} ends the interaction;
the benchmark's verifier determines the score.

\Needspace{20\baselineskip}
\subsection{Hosted and Local Evaluation}

Listing~\ref{lst:evaluation-command} runs the Codex template on the
representative OSWorld set using Modal. The command specifies the
submission directory, benchmark, and number of parallel evaluations.
Omitting \texttt{--remote} runs both the agent and environment VMs on the
evaluator host. For local desktop evaluation, this must be a Linux host
with the VM runtime and hardware required by the benchmark, as well as
model-serving resources when using open-weight models.

\begin{agentlisting}{Evaluating a supplied agent template.\label{lst:evaluation-command}}
cua-speedrun run --remote \
    --submission templates/codex_cli \
    --benchmark benchmarks/osworld-energy50-representative \
    --agents-per-evaluation 1 --parallel-evaluations 8
\end{agentlisting}

Users can also submit agents through the dashboard. For each evaluation,
we record the benchmark, hardware, evaluation settings, and task seeds,
along with task scores and timings. Action logs and screenshots allow
users to inspect how the agent completed or failed each task.

\FloatBarrier
\section{Interaction and Token Measurements}
\label{app:measurements}

Alongside performance, time, and cost, we measure how often an agent calls
its model, how many actions it executes, and how many tokens it generates.
These measurements support the analysis of agent speed in
Sec.~\ref{sec:results}. The full configuration tables average over all
tasks, including unsuccessful attempts.

\paragraph{Aggregate metrics.}
For an agent $\pi$ evaluated on $K$ tasks, the mean verifier score $P$,
mean task time $T$, and mean model cost $C$ are
\begin{equation}
    P(\pi)=\frac{1}{K}\sum_{i=1}^{K}r_i, \qquad
    T(\pi)=\frac{1}{K}\sum_{i=1}^{K}t_i, \qquad
    C(\pi)=\frac{1}{K}\sum_{i=1}^{K}c_i,
\end{equation}
where $r_i$ is the verifier score defined in
Appendix~\ref{app:task-definition}, and $t_i$ and $c_i$ are the execution
time and model cost of task $i$. The mean score retains partial credit; we
also report the fraction of exactly completed tasks and the median time on
those tasks.

\paragraph{Agent time, environment time, and requested waits.}
\label{app:timing-components}
We separate total task time into agent time and environment time.
For task $i$, environment time $e_i$ sums the intervals recorded by the
environment server for executing actions and returning observations.
This includes explicit waits requested by the agent; we report their
requested durations separately and subtract them when reporting
environment time excluding agent waits. Agent time is the remainder,
$t_i-e_i$, which includes model requests and agent-side processing.

\subsection{Model Calls, Action Batches, and Individual Actions}

\paragraph{Model responses.}
We count each completed model response, including responses without a
computer action. A response can contain several tool calls, each of
which may execute several actions. We therefore count model responses
and desktop actions separately.

\paragraph{Action batches and individual actions.}
An agent can execute several actions in one request. For example,
clicking a field, typing text, and pressing Enter in one request counts
as one batch and three actions. Issuing these actions separately counts
as three batches and three actions. A \emph{step} denotes one request
containing keyboard, mouse, or wait actions. For task $i$ with $S_i$ batches, let
$\mathcal{A}_{ij}$ be the action list in request $j$. The total number of
individual actions is $A_i=\sum_{j=1}^{S_i}|\mathcal{A}_{ij}|$.

We count actions as submitted by the agent: typing a string, pressing a
key combination, or requesting a wait each counts as one action.

\subsection{Token Accounting}
\label{app:token-accounting}

We count every generated token once per response, including reasoning
and other output, using the provider's reported usage. We also report
provider-supplied reasoning-token counts separately.

To compare token rates, we divide total generated tokens by total agent
time: $\sum_i G_i / \sum_i(t_i-e_i)$, where $G_i$ is the generated-token
count for task $i$. This \emph{effective token rate} includes the
time spent on model requests and agent-side processing.

\subsection{Full Interaction and Token Comparisons}

Tables~\ref{tab:interaction-osworld} and~\ref{tab:interaction-osworld2}
report mean batches, individual actions, model responses, and generated
tokens for the configurations in Appendix~\ref{app:full-results}.
Configuration IDs match the frontier plots and result tables.

\begingroup\footnotesize
\setlength{\tabcolsep}{4pt}
\begin{longtable}{rp{0.39\linewidth}rrrr}
\caption{\textbf{Interaction and token counts on OSWorld.} Means per task, including unsuccessful tasks. IDs match the OSWorld configuration tables in Appendix~\ref{app:full-results}. Tokens include reasoning and other output. Astra (low) uses five-seed means for each I/O setting.}\label{tab:interaction-osworld}\\
\toprule
ID & Configuration & Batches & Actions & Responses & Tokens \\
\midrule
\endfirsthead
\multicolumn{6}{l}{\tablename~\thetable{}: OSWorld interaction counts, continued.}\\
\toprule
ID & Configuration & Batches & Actions & Responses & Tokens \\
\midrule
\endhead
\midrule\multicolumn{6}{r}{Continued on next page}\\
\endfoot
\bottomrule
\endlastfoot
1 & GPT-6 Astra / xhigh & 7.14 & 19.90 & -- & 2,060 \\
2 & Gemini 3.8 Flash / low & 16.60 & 20.02 & 18.52 & 1,564 \\
3 & Claude Opus 5 / high & 19.92 & 26.92 & 20.62 & 3,735 \\
4 & Gemini 3.8 Flash / medium & 27.16 & 34.70 & 29.10 & 5,365 \\
5 & GPT-6 Astra / medium & 6.30 & 19.26 & -- & 1,366 \\
6 & GPT-6 Astra / high & 6.52 & 19.76 & -- & 1,593 \\
7 & GPT-6 Astra / low / fast & 8.20 & 27.10 & -- & 1,328 \\
8 & Claude Sonnet 5 / high & 20.24 & 29.20 & 22.28 & 4,840 \\
9 & Gemini 3.7 Flash / medium & 19.44 & 23.34 & 21.36 & 3,594 \\
10 & Gemini 3.7 Flash / high & 24.26 & 28.30 & 26.20 & 5,731 \\
11 & Kimi K3 / max / batched & 11.70 & 53.58 & 11.68 & 10,435 \\
12 & Claude Opus 5 / low & 14.04 & 23.80 & 16.08 & 1,795 \\
13 & GPT-6 Astra / low & 5.86 & 19.10 & -- & 1,579 \\
14 & Gemini 3.7 Flash / low & 18.24 & 20.88 & 20.14 & 2,944 \\
15 & Gemini 3.8 Flash / high & 37.18 & 51.46 & 39.08 & 8,789 \\
16 & GPT-5.6 Luna / medium / API & 14.42 & 48.88 & 15.34 & 1,751 \\
17 & GPT-5.6 Sol / xhigh & 12.66 & 36.78 & 13.62 & 2,402 \\
18 & Kimi K3 / high / single & 8.92 & 113.60 & 10.70 & 7,567 \\
19 & GPT-5.6 Sol / medium & 12.72 & 42.64 & 13.68 & 2,717 \\
20 & GPT-5.6 Luna / high / Codex & 15.58 & 39.80 & -- & 8,637 \\
21 & Kimi K3 / high / batched & 8.76 & 60.20 & 9.20 & 7,125 \\
22 & Muse Spark 1.1 / xhigh & 69.82 & 145.56 & 26.02 & 10,701 \\
23 & GPT-5.6 Luna / xhigh / Codex & 13.88 & 33.10 & -- & 8,160 \\
24 & Muse Spark 1.3 / xhigh & 56.58 & 86.72 & 30.62 & 11,505 \\
25 & Muse Spark 1.3 / medium & 41.54 & 69.32 & 26.10 & 8,952 \\
26 & Claude Sonnet 5 / low & 21.24 & 29.50 & 24.16 & 3,468 \\
27 & GPT-5.6 Luna / low / Codex & 7.22 & 26.44 & -- & 3,191 \\
28 & MiniMax M3 / thinking\_off & 34.32 & 51.10 & 35.14 & -- \\
29 & GPT-5.6 Luna / high / API & 18.28 & 53.88 & 19.24 & 3,358 \\
30 & Kimi K3 / low / batched & 5.74 & 30.00 & 6.90 & 1,865 \\
31 & Muse Spark 1.3 / high & 51.30 & 84.78 & 28.98 & 10,538 \\
32 & Kimi K3 / low / single & 8.12 & 55.10 & 9.62 & 3,171 \\
33 & MiniMax M3 / thinking\_on & 35.00 & 55.30 & 35.94 & -- \\
34 & Muse Spark 1.1 / medium & 54.36 & 114.00 & 22.82 & 7,352 \\
35 & Muse Spark 1.1 / high & 69.90 & 126.88 & 25.66 & 10,132 \\
36 & Muse Spark 1.1 / low & 58.48 & 149.86 & 22.12 & 6,526 \\
37 & Muse Spark 1.2 / xhigh & 66.54 & 200.52 & 43.08 & 17,198 \\
38 & GPT-5.6 Luna / medium / Codex & 9.50 & 25.64 & -- & 4,524 \\
39 & Muse Spark 1.1 / minimal & 53.44 & 131.68 & 20.62 & 5,957 \\
40 & Muse Spark 1.3 / low & 40.50 & 71.82 & 23.54 & 6,747 \\
41 & GPT-5.6 Sol / low & 11.20 & 45.34 & 12.16 & 1,236 \\
42 & Muse Spark 1.2 / low & 31.12 & 137.82 & 23.94 & 4,723 \\
43 & Muse Spark 1.2 / medium & 44.20 & 152.84 & 28.34 & 7,789 \\
44 & GPT-5.6 Luna / low / API & 10.12 & 37.66 & 11.06 & 1,063 \\
45 & Muse Spark 1.3 / minimal & 32.04 & 59.52 & 19.76 & 3,792 \\
46 & Muse Spark 1.2 / high & 45.72 & 145.58 & 33.74 & 9,551 \\
47 & Gemini 3 Flash Preview / high & 33.28 & 34.56 & 37.86 & 5,167 \\
48 & GLM-5V Turbo / thinking\_off & 16.86 & 21.52 & 17.54 & 8,150 \\
49 & Gemini 3 Flash Preview / medium & 28.26 & 30.68 & 32.52 & 4,125 \\
50 & GLM-5V Turbo / thinking\_on & 17.86 & 19.24 & 18.62 & 8,563 \\
51 & Muse Spark 1.2 / minimal & 23.20 & 87.38 & 18.72 & 2,379 \\
52 & Gemini 3 Flash Preview / low & 39.52 & 40.26 & 56.02 & 1,598 \\
53 & Yutori n2 / none & 18.74 & 56.54 & 20.90 & 3,652 \\
54 & Yutori n2 / low & 22.96 & 58.72 & 25.32 & 12,188 \\
55 & Yutori n2 / medium & 19.06 & 55.14 & 21.12 & 7,562 \\
56 & Yutori n2 / xhigh & 17.56 & 48.16 & 19.72 & 7,499 \\
\end{longtable}
\endgroup

\begingroup\footnotesize
\setlength{\tabcolsep}{4pt}
\begin{longtable}{rp{0.39\linewidth}rrrr}
\caption{\textbf{Interaction and token counts on OSWorld2.} Means per task, including unsuccessful tasks. IDs match the OSWorld2 configuration tables in Appendix~\ref{app:full-results}. Tokens include reasoning and other output.}\label{tab:interaction-osworld2}\\
\toprule
ID & Configuration & Batches & Actions & Responses & Tokens \\
\midrule
\endfirsthead
\multicolumn{6}{l}{\tablename~\thetable{}: OSWorld2 interaction counts, continued.}\\
\toprule
ID & Configuration & Batches & Actions & Responses & Tokens \\
\midrule
\endhead
\midrule\multicolumn{6}{r}{Continued on next page}\\
\endfoot
\bottomrule
\endlastfoot
1 & GPT-6 Astra / xhigh & 52.81 & 194.10 & -- & 18,332 \\
2 & GPT-6 Astra / high & 51.29 & 149.02 & -- & 14,146 \\
3 & GPT-6 Astra / low & 42.58 & 151.40 & -- & 10,190 \\
4 & GPT-6 Astra / medium & 45.46 & 151.08 & -- & 11,382 \\
5 & Gemini 3.8 Flash / high & 216.54 & 331.88 & 218.48 & 72,149 \\
6 & Gemini 3.8 Flash / medium & 167.79 & 252.65 & 169.65 & 49,532 \\
7 & Claude Opus 5 / low & 146.52 & 316.17 & 158.29 & 30,820 \\
8 & Claude Sonnet 5 / low & 237.25 & 378.54 & 253.50 & 57,018 \\
9 & Gemini 3.8 Flash / low & 127.04 & 200.75 & 128.83 & 19,840 \\
10 & Muse Spark 1.3 / minimal & 115.35 & 488.06 & 66.02 & 16,150 \\
11 & Muse Spark 1.3 / low & 130.50 & 655.08 & 80.73 & 25,131 \\
12 & Muse Spark 1.3 / medium & 149.81 & 832.44 & 89.75 & 41,149 \\
13 & Muse Spark 1.3 / xhigh & 147.29 & 832.75 & 93.96 & 52,628 \\
14 & Muse Spark 1.3 / high & 151.21 & 754.98 & 89.83 & 44,241 \\
15 & Kimi K3 / low & 47.25 & 313.75 & 49.17 & 21,852 \\
16 & MiniMax M3 / thinking\_on & 97.56 & 407.33 & 98.62 & -- \\
17 & MiniMax M3 / thinking\_off & 98.54 & 368.62 & 98.65 & -- \\
18 & GLM-5V Turbo / thinking\_off & 49.15 & 79.35 & 48.38 & 28,558 \\
19 & GLM-5V Turbo / thinking\_on & 47.75 & 91.75 & 47.12 & 28,385 \\
20 & Kimi K3 / high & 58.77 & 350.06 & 59.04 & 81,494 \\
21 & Kimi K3 / max & 66.38 & 386.56 & 60.79 & 123,603 \\
\end{longtable}
\endgroup

\clearpage
\begin{figure}[!t]
  \centering
  \includegraphics[width=\linewidth]{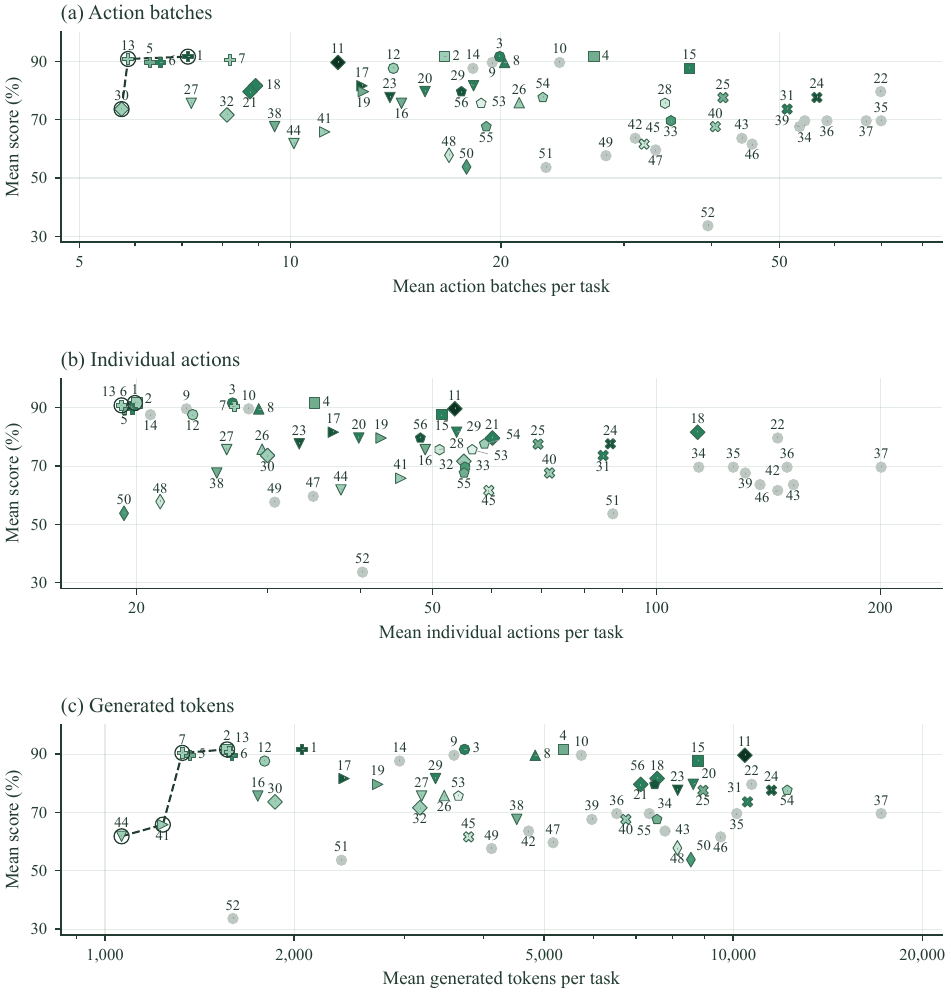}
  \caption{\textbf{Performance--interaction and performance--token frontiers on OSWorld.}
  Mean score against mean action batches, individual actions, and generated
  tokens per task. Tokens include reasoning and other output. Panels (a)--(b)
  show all 56 configurations; panel (c) shows the 54 with token counts.
  Dashed lines connect the circled frontier points. IDs match the
  configuration tables in Appendix~\ref{app:full-results}; darker green
  indicates greater reasoning effort. Astra (low) uses five-seed means for
  each I/O setting.}
  \label{fig:interaction-frontiers-osworld}
\end{figure}
\FloatBarrier

\subsection{Performance--Interaction and Performance--Token Frontiers}
\label{app:interaction-frontiers}

\begin{figure}[H]
  \centering
  \includegraphics[width=\linewidth]{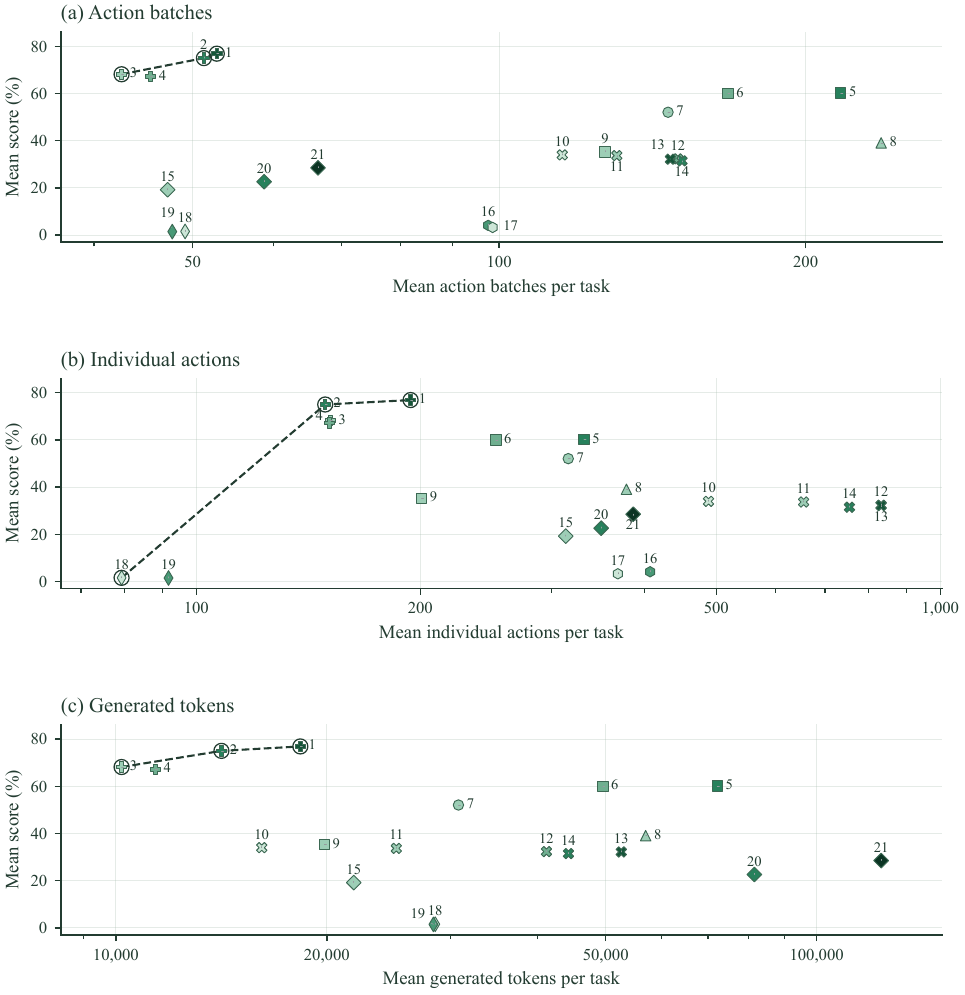}
  \caption{\textbf{Performance--interaction and performance--token frontiers on OSWorld2.}
  Mean score against mean action batches, individual actions, and generated
  tokens per task. Panels (a)--(b) show all 21 configurations; panel (c)
  shows the 19 with token counts. IDs match the OSWorld2 configuration
  table in Appendix~\ref{app:full-results}. Other plotting conventions follow
  Figure~\ref{fig:interaction-frontiers-osworld}.}
  \label{fig:interaction-frontiers-osworld2}
\end{figure}

Figures~\ref{fig:interaction-frontiers-osworld}
and~\ref{fig:interaction-frontiers-osworld2} compare performance with the
number of action batches, individual actions, and generated tokens on
OSWorld and OSWorld2. Circled points mark the Pareto frontier for each
metric. We use the same
trajectories as the time and cost comparisons and average over all tasks.

\clearpage
\FloatBarrier

\clearpage
\section{Full Evaluation Results}
\label{app:full-results}

\subsection{Model, Harness, and Reasoning-Effort Configurations}
\label{app:configurations}

Tables~\ref{tab:configurations-osworld-1}--\ref{tab:configurations-osworld2}
report mean score, time, cost, and token use for each model, harness,
and reasoning setting. Astra (low) uses five-seed means for each I/O setting.
The configuration IDs identify the corresponding points in the figures.

We evaluate Luna through both the direct API and Codex, and Kimi with
single-call and batched-call harnesses. We use the Yutori cookbook.

\begin{table}[!ht]
\centering\scriptsize
\setlength{\tabcolsep}{3pt}
\begin{tabular}{rlrrrr}
\toprule
ID & Configuration & Score (\%) & Time (s) & Cost (\$) & Tokens \\
\midrule
\textbf{1} & GPT-6 Astra / xhigh & 91.6 & 126.8 & 0.708 & 2,060 \\
\textbf{2} & Gemini 3.8 Flash / low & 91.6 & 127.1 & 0.108 & 1,564 \\
3 & Claude Opus 5 / high & 91.6 & 136.0 & 0.661 & 3,735 \\
4 & Gemini 3.8 Flash / medium & 91.6 & 250.1 & 0.221 & 5,365 \\
5 & GPT-6 Astra / medium & 89.6 & 90.2 & 0.600 & 1,366 \\
6 & GPT-6 Astra / high & 89.6 & 100.8 & 0.636 & 1,593 \\
7 & GPT-6 Astra / low / fast & 90.4 & 99.0 & 0.908 & 1,328 \\
8 & Claude Sonnet 5 / high & 89.6 & 140.8 & 0.309 & 4,840 \\
9 & Gemini 3.7 Flash / medium & 89.6 & 198.9 & 0.944 & 3,594 \\
10 & Gemini 3.7 Flash / high & 89.6 & 254.3 & 1.412 & 5,731 \\
11 & Kimi K3 / max / batched & 89.6 & 443.7 & 0.503 & 10,435 \\
\textbf{12} & Claude Opus 5 / low & 87.6 & 85.9 & 0.390 & 1,795 \\
\textbf{13} & GPT-6 Astra / low & 90.8 & 89.5 & 0.531 & 1,579 \\
14 & Gemini 3.7 Flash / low & 87.6 & 207.3 & 0.926 & 2,944 \\
15 & Gemini 3.8 Flash / high & 87.6 & 347.7 & 0.347 & 8,789 \\
\textbf{16} & GPT-5.6 Luna / medium / API & 75.6 & 118.9 & 0.019 & 1,751 \\
17 & GPT-5.6 Sol / xhigh & 81.6 & 108.3 & 0.527 & 2,402 \\
18 & Kimi K3 / high / single & 81.6 & 305.1 & 0.414 & 7,567 \\
19 & GPT-5.6 Sol / medium & 79.6 & 121.3 & 0.539 & 2,717 \\
20 & GPT-5.6 Luna / high / Codex & 79.6 & 324.8 & 0.076 & 8,637 \\
21 & Kimi K3 / high / batched & 79.6 & 358.1 & 0.382 & 7,125 \\
22 & Muse Spark 1.1 / xhigh & 79.6 & 417.2 & 1.685 & 10,701 \\
23 & GPT-5.6 Luna / xhigh / Codex & 77.6 & 297.2 & 0.063 & 8,160 \\
24 & Muse Spark 1.3 / xhigh & 77.6 & 612.1 & 0.161 & 11,505 \\
25 & Muse Spark 1.3 / medium & 77.5 & 397.6 & 0.126 & 8,952 \\
26 & Claude Sonnet 5 / low & 75.8 & 129.4 & 0.294 & 3,468 \\
\bottomrule
\end{tabular}
\caption{OSWorld configurations. Bold IDs mark the joint performance--time--cost frontier. Tokens include reasoning and other generated output.}
\label{tab:configurations-osworld-1}
\end{table}

\begin{table}[!t]
\centering\scriptsize
\setlength{\tabcolsep}{3pt}
\begin{tabular}{rlrrrr}
\toprule
ID & Configuration & Score (\%) & Time (s) & Cost (\$) & Tokens \\
\midrule
27 & GPT-5.6 Luna / low / Codex & 75.6 & 144.8 & 0.025 & 3,191 \\
28 & MiniMax M3 / thinking\_off & 75.6 & 253.8 & -- & -- \\
\textbf{29} & GPT-5.6 Luna / high / API & 81.6 & 160.5 & 0.029 & 3,358 \\
30 & Kimi K3 / low / batched & 73.6 & 149.5 & 0.174 & 1,865 \\
31 & Muse Spark 1.3 / high & 73.6 & 506.9 & 0.148 & 10,538 \\
32 & Kimi K3 / low / single & 71.6 & 176.6 & 0.280 & 3,171 \\
33 & MiniMax M3 / thinking\_on & 69.6 & 345.1 & -- & -- \\
34 & Muse Spark 1.1 / medium & 69.6 & 359.8 & 1.379 & 7,352 \\
35 & Muse Spark 1.1 / high & 69.6 & 368.5 & 1.706 & 10,132 \\
36 & Muse Spark 1.1 / low & 69.6 & 370.4 & 1.312 & 6,526 \\
37 & Muse Spark 1.2 / xhigh & 69.6 & 720.9 & 0.244 & 17,198 \\
38 & GPT-5.6 Luna / medium / Codex & 67.6 & 188.8 & 0.035 & 4,524 \\
39 & Muse Spark 1.1 / minimal & 67.6 & 318.5 & 1.184 & 5,957 \\
40 & Muse Spark 1.3 / low & 67.6 & 361.0 & 0.107 & 6,747 \\
\textbf{41} & GPT-5.6 Sol / low & 65.8 & 100.1 & 0.319 & 1,236 \\
42 & Muse Spark 1.2 / low & 63.6 & 363.6 & 0.105 & 4,723 \\
43 & Muse Spark 1.2 / medium & 63.6 & 466.2 & 0.136 & 7,789 \\
\textbf{44} & GPT-5.6 Luna / low / API & 61.8 & 73.6 & 0.011 & 1,063 \\
45 & Muse Spark 1.3 / minimal & 61.6 & 264.4 & 0.079 & 3,792 \\
46 & Muse Spark 1.2 / high & 61.6 & 669.8 & 0.174 & 9,551 \\
47 & Gemini 3 Flash Preview / high & 59.6 & 337.2 & 0.809 & 5,167 \\
48 & GLM-5V Turbo / thinking\_off & 57.8 & 752.4 & 0.220 & 8,150 \\
49 & Gemini 3 Flash Preview / medium & 57.6 & 268.0 & 0.592 & 4,125 \\
50 & GLM-5V Turbo / thinking\_on & 53.8 & 359.2 & 0.235 & 8,563 \\
51 & Muse Spark 1.2 / minimal & 53.6 & 248.8 & 0.070 & 2,379 \\
52 & Gemini 3 Flash Preview / low & 33.6 & 492.2 & 1.401 & 1,598 \\
53 & Yutori n2 / none & 75.6 & 298.5 & 0.081 & 3,652 \\
54 & Yutori n2 / low & 77.6 & 331.0 & 0.157 & 12,188 \\
55 & Yutori n2 / medium & 67.6 & 265.4 & 0.110 & 7,562 \\
56 & Yutori n2 / xhigh & 79.6 & 225.0 & 0.104 & 7,499 \\
\bottomrule
\end{tabular}
\caption{OSWorld configurations, continued. Bold IDs mark the joint performance--time--cost frontier. Tokens include reasoning and other generated output.}
\label{tab:configurations-osworld-2}
\end{table}

\begin{table}[!t]
\centering\scriptsize
\setlength{\tabcolsep}{3pt}
\begin{tabular}{rlrrrr}
\toprule
ID & Configuration & Score (\%) & Time (s) & Cost (\$) & Tokens \\
\midrule
\textbf{1} & GPT-6 Astra / xhigh & 76.9 & 1237.3 & 8.175 & 18,332 \\
\textbf{2} & GPT-6 Astra / high & 75.0 & 914.0 & 7.391 & 14,146 \\
\textbf{3} & GPT-6 Astra / low & 68.2 & 904.2 & 6.336 & 10,190 \\
\textbf{4} & GPT-6 Astra / medium & 67.2 & 828.1 & 6.519 & 11,382 \\
\textbf{5} & Gemini 3.8 Flash / high & 60.2 & 3058.8 & 5.019 & 72,149 \\
\textbf{6} & Gemini 3.8 Flash / medium & 59.9 & 2051.8 & 2.915 & 49,532 \\
7 & Claude Opus 5 / low & 52.1 & 1385.7 & 10.006 & 30,820 \\
8 & Claude Sonnet 5 / low & 39.1 & 1993.6 & 8.863 & 57,018 \\
\textbf{9} & Gemini 3.8 Flash / low & 35.3 & 1386.7 & 1.875 & 19,840 \\
\textbf{10} & Muse Spark 1.3 / minimal & 34.0 & 1516.0 & 0.390 & 16,150 \\
11 & Muse Spark 1.3 / low & 33.7 & 2280.8 & 0.503 & 25,131 \\
12 & Muse Spark 1.3 / medium & 32.3 & 2292.1 & 0.588 & 41,149 \\
13 & Muse Spark 1.3 / xhigh & 32.2 & 2261.0 & 0.640 & 52,628 \\
14 & Muse Spark 1.3 / high & 31.5 & 2179.3 & 0.599 & 44,241 \\
15 & Kimi K3 / low & 19.2 & 1860.9 & 2.088 & 21,852 \\
16 & MiniMax M3 / thinking\_on & 4.1 & 1673.2 & -- & -- \\
17 & MiniMax M3 / thinking\_off & 3.2 & 1446.9 & -- & -- \\
18 & GLM-5V Turbo / thinking\_off & 1.5 & 1979.4 & 0.739 & 28,558 \\
19 & GLM-5V Turbo / thinking\_on & 1.5 & 2641.9 & 0.725 & 28,385 \\
20 & Kimi K3 / high & 22.6 & 3733.9 & 4.517 & 81,494 \\
21 & Kimi K3 / max & 28.5 & 4825.1 & 5.625 & 123,603 \\
\bottomrule
\end{tabular}
\caption{OSWorld2 configurations. Bold IDs mark the joint performance--time--cost frontier. Tokens include reasoning and other generated output.}
\label{tab:configurations-osworld2}
\end{table}

\clearpage
\begin{figure}[!t]
  \centering
  \includegraphics[width=0.92\linewidth]{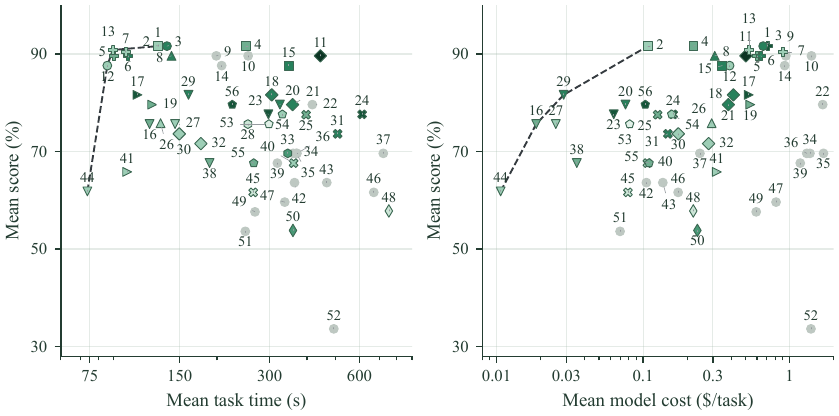}
  \caption{\textbf{Full OSWorld performance--time and performance--cost comparisons.}
  All 56 configurations appear in the time comparison; the cost comparison
  uses configurations with reported or estimated costs. IDs identify the
  model, reasoning effort, and harness in Tables~\ref{tab:configurations-osworld-1}--\ref{tab:configurations-osworld-2}.
  Dashed lines connect nondominated configurations for each pair of axes.
  Scores, times, and costs are means across the same 50 tasks, including unsuccessful tasks.}
  \label{fig:full-osworld}
\end{figure}

\begin{figure}[!t]
  \centering
  \includegraphics[width=0.92\linewidth]{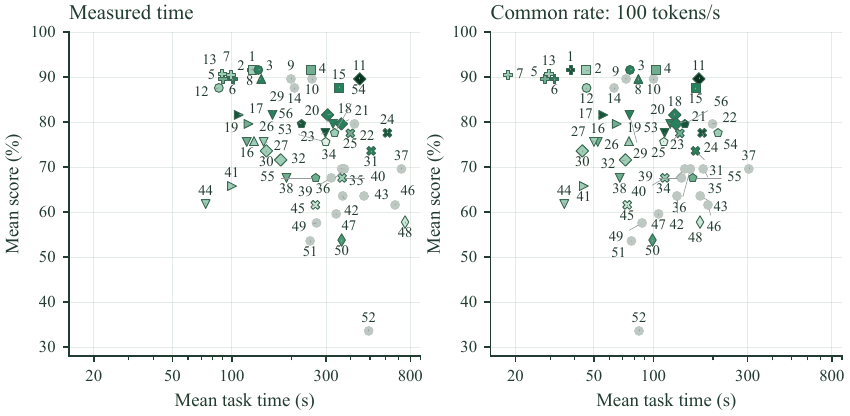}
  \caption{\textbf{Full OSWorld comparison at a common token rate.}
  Both plots use the same 54 configurations with token counts.
  The right plot replaces measured mean task time with mean recorded
  environment time plus generated tokens divided by 100 tokens/s.
  Tokens include reasoning and other output; scores and trajectories are
  unchanged. Point IDs refer to the OSWorld configuration tables.}
  \label{fig:full-normalized}
\end{figure}
\FloatBarrier

\subsection{OSWorld: Performance--Time and Performance--Cost Frontiers}
\label{app:osworld-frontiers}

Figure~\ref{fig:full-osworld} shows the full OSWorld comparison from
Figure~\ref{fig:results-overview}, with configuration IDs for all 56 points.

\paragraph{A common token-generation rate.}
Figure~\ref{fig:full-normalized} compares measured task time with an
estimate that assigns every model the same generation rate. We compute
this estimate as recorded environment time plus generated tokens divided
by 100 tokens/s, using the same trajectories and scores. Opus 5 (low)
scores 87.6\% in 85.9\,s, while Astra (low) scores 90.8\% in 89.5\,s.
Opus generates 1,795 tokens per task compared with 1,579 for Astra.
With generation time assigned at the common rate,
Astra is faster. Environment time includes agent-requested waits
(Appendix~\ref{app:timing-components}).

\clearpage
\begin{figure}[!t]
  \centering
  \includegraphics[width=0.92\linewidth]{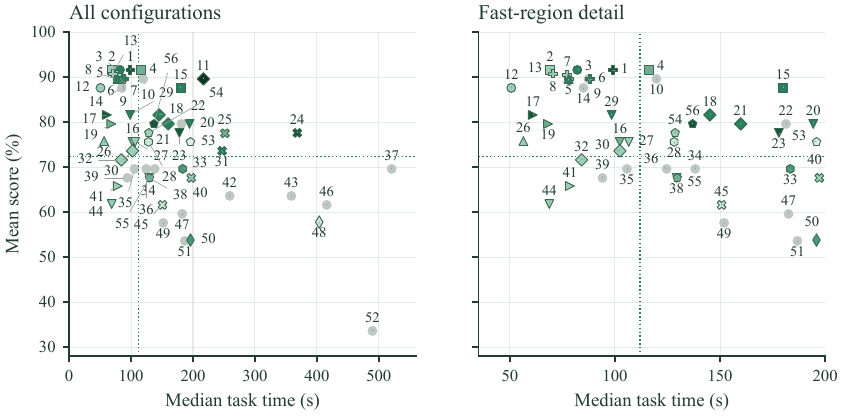}
  \caption{\textbf{Full OSWorld median-time view and fast-region detail.}
  Mean verifier score versus median task time for all 56 configurations.
  For Astra (low), the median task times are averaged across five seeds.
  The right plot enlarges the region containing the fastest configurations;
  point IDs match the full OSWorld tables. Green dotted lines show the
  published human references of 72.36\% and 111.94\,s from the original
  369-task OSWorld benchmark~\citep{xie2024osworldbenchmarkingmultimodalagents}.}
  \label{fig:full-median}
\end{figure}

\begin{figure}[!t]
  \centering
  \includegraphics[width=0.92\linewidth]{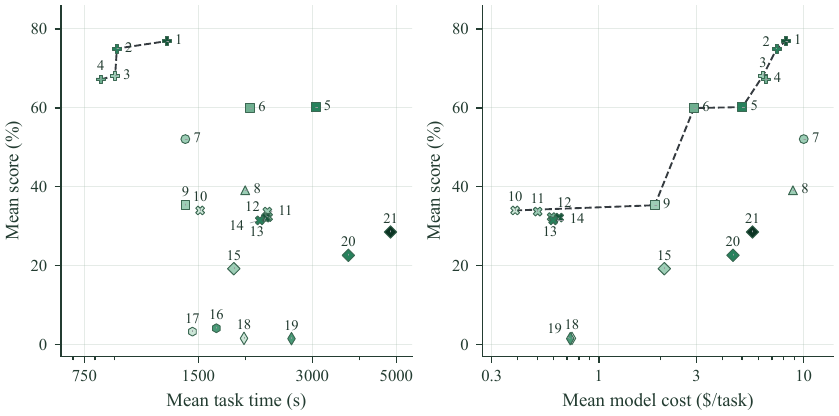}
  \caption{\textbf{Full OSWorld2 performance--time and performance--cost comparisons.}
  All 21 configurations are evaluated on the same 52 tasks.
  Point IDs refer to the OSWorld2 configuration table. Marker shapes follow
  Figure~\ref{fig:results-overview}; darker green indicates greater reasoning
  effort.}
  \label{fig:full-osworld2}
\end{figure}
\FloatBarrier

\paragraph{Median task time.}
Figure~\ref{fig:full-median} reports median task time, which summarizes
the time for a typical task.

\subsection{OSWorld2: Performance--Time and Performance--Cost Frontiers}
\label{app:osworld2-frontiers}

Figure~\ref{fig:full-osworld2} compares all 21 OSWorld2 configurations.
Astra defines the entire time frontier, while Gemini 3.8 Flash and Muse
Spark 1.3 offer lower-cost choices on the cost frontier.

\clearpage
\begin{figure}[!t]
  \centering
  \includegraphics[width=\linewidth]{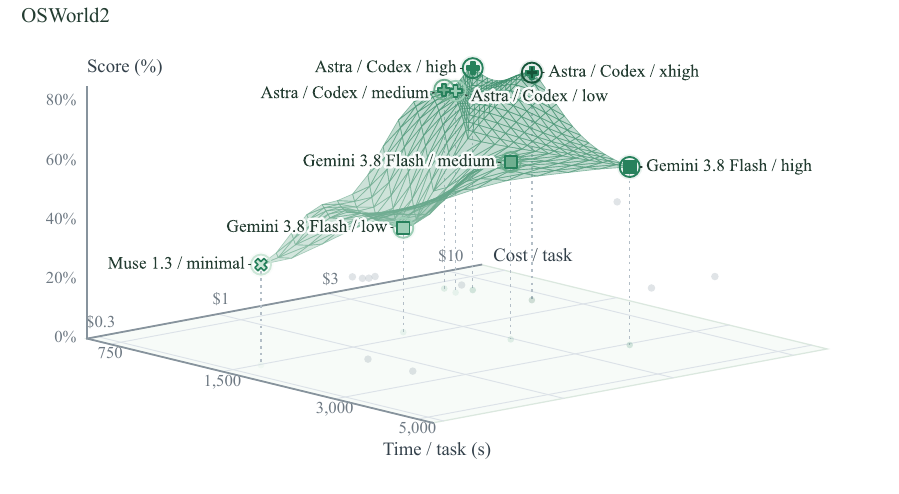}
  \caption{\textbf{Joint performance--time--cost frontier on OSWorld2.}
  Circled points are the evaluated frontier configurations. Other configurations
  appear in gray. Time and cost axes are logarithmic. Marker shapes follow
  Figure~\ref{fig:results-overview}; darker green indicates greater reasoning effort.}
  \label{fig:joint-osworld2}
\end{figure}
\FloatBarrier

\subsection{OSWorld2: Joint Performance--Time--Cost Frontier}
\label{app:osworld2-joint}

Figure~\ref{fig:joint-osworld2} extends the joint frontier in
Figure~\ref{fig:results-joint-frontier} to OSWorld2. A configuration is on
this frontier if no other evaluated configuration improves one of the
three metrics without worsening another. It includes Astra at all four
reasoning settings, Gemini 3.8 Flash at all three settings, and Muse
Spark 1.3 at minimal effort, allowing a choice based on the desired
score and the available time and cost budgets.

\clearpage
\begin{figure}[!t]
  \centering
  \includegraphics[width=0.88\linewidth]{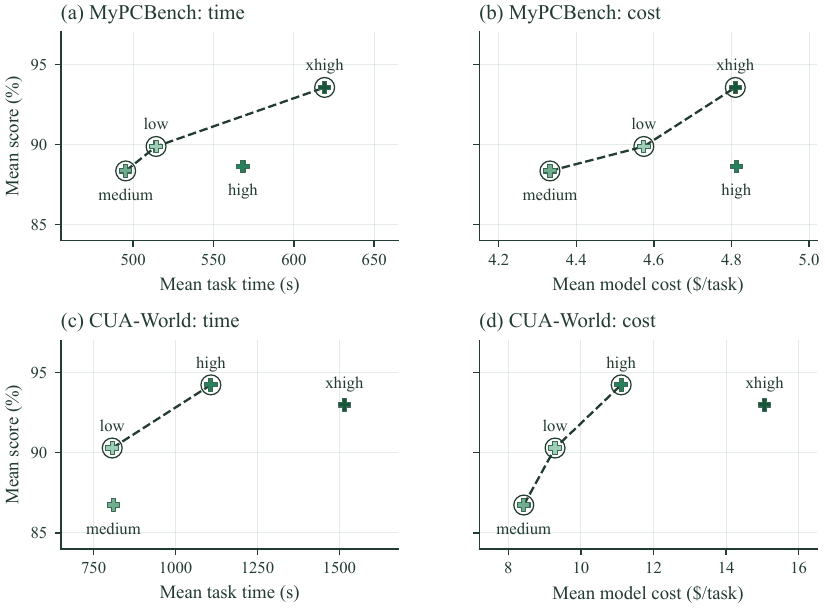}
  \caption{\textbf{Performance--time and performance--cost frontiers on MyPCBench and CUA-World.}
  GPT-6 Astra through Codex at four reasoning settings. Each point averages
  the same 38 MyPCBench tasks (top) or 26 CUA-World tasks (bottom).
  Darker green indicates greater reasoning effort. Dashed lines connect
  the observed Pareto-frontier points, which are circled.}
  \label{fig:additional-benchmarks}
\end{figure}

\begin{table}[!t]
  \centering
  \small
  \begin{tabular}{lrrrrrr}
    \toprule
    & \multicolumn{3}{c}{MyPCBench (38 tasks)}
    & \multicolumn{3}{c}{CUA-World (26 tasks)} \\
    \cmidrule(lr){2-4}\cmidrule(lr){5-7}
    Effort & Score (\%) & Time (s) & Cost (\$)
           & Score (\%) & Time (s) & Cost (\$) \\
    \midrule
    Low    & 89.9 & 514 & 4.57 & 90.3 & \textbf{807} & 9.29 \\
    Medium & 88.3 & \textbf{495} & \textbf{4.33} & 86.7 & 810 & \textbf{8.42} \\
    High   & 88.6 & 568 & 4.81 & \textbf{94.2} & 1,108 & 11.11 \\
    Xhigh  & \textbf{93.6} & 619 & 4.81 & 93.0 & 1,515 & 15.06 \\
    \bottomrule
  \end{tabular}
  \caption{\textbf{Reasoning effort on MyPCBench and CUA-World.}
  GPT-6 Astra through Codex, reporting mean partial-credit score, time,
  and model cost per task. Bold indicates the highest score,
  lowest time, or lowest cost within each benchmark.}
  \label{tab:additional-benchmarks}
\end{table}
\FloatBarrier

\subsection{Reasoning Effort on MyPCBench}
\label{app:additional-benchmarks}

Figure~\ref{fig:additional-benchmarks} and
Table~\ref{tab:additional-benchmarks} compare GPT-6 Astra through Codex at
four reasoning settings on MyPCBench. Xhigh achieves the highest score,
93.6\%, while medium has the lowest mean time and cost, 495\,s and
\$4.33 per task. We evaluate the 38-task representative
set from Sec.~\ref{sec:representative-tasks}, with a limit of 100 action
batches. All reported means include
unsuccessful tasks.

\subsection{Reasoning Effort on CUA-World}

Figure~\ref{fig:additional-benchmarks} and
Table~\ref{tab:additional-benchmarks} show a different outcome on
CUA-World. High effort scores 94.2\%, compared
with 93.0\% at xhigh, while taking less time (1,108 vs.\ 1,515\,s)
and costing less (\$11.11 vs.\ \$15.06 per task).
All four settings use Astra through Codex on the same 26 long-horizon
tasks, with a limit of 500 action batches.

\clearpage
\section{Run-to-Run Variation}
\label{app:run-variation}

We repeat the evaluation across five seeds to measure variation in
scores, task times, and action sequences. We also compare scores on
individual tasks to determine whether the same tasks succeed across
runs. Appendix~\ref{app:fast-io-consistency} reports the corresponding
timing results for Fast I/O.

\subsection{Five-Seed Evaluation Setup}

We evaluate GPT-6 Astra at low reasoning effort using Codex on the
same 50 OSWorld tasks. Each run uses the same agent prompt and a
500-step limit. Only the task seed changes.

\begin{table}[!t]
  \centering
  \footnotesize
  \setlength{\tabcolsep}{4pt}
  \begin{tabular}{lrrrrrrr}
    \toprule
    Run & Score (\%) & Time (s) & Agent (s) & Steps & Env. (s) & Waits (s) & Env.$-$waits (s) \\
    \midrule
    1 & 87.62 & 86.11 & 72.90 & 5.90 & 13.209 & 1.328 & 11.881 \\
    2 & 91.62 & 89.56 & 75.31 & 5.82 & 14.248 & 1.082 & 13.166 \\
    3 & 91.62 & 90.13 & 76.52 & 5.88 & 13.606 & 1.040 & 12.566 \\
    4 & 91.62 & 89.81 & 75.78 & 5.78 & 14.033 & 1.152 & 12.881 \\
    5 & 91.62 & 92.10 & 78.19 & 5.94 & 13.914 & 1.012 & 12.902 \\
    Mean & 90.82 & 89.54 & 75.74 & 5.86 & 13.802 & 1.123 & 12.679 \\
    SD & 1.79 & 2.16 & 1.93 & 0.06 & 0.405 & 0.126 & 0.494 \\
    \bottomrule
  \end{tabular}
  \caption{\textbf{Variation across five seeds for GPT-6 Astra at low reasoning effort.}
  Each row averages all 50 OSWorld tasks. SD is the sample standard
  deviation of these means across the five runs;
  score SD is in percentage points. Steps count action batches, and
  waits are explicitly requested by the agent.}
  \label{tab:run-variation}
\end{table}

\subsection{Scores and Times Across Seeds}

Table~\ref{tab:run-variation} shows that both score and task time vary
little across seeds. Mean task time is $89.54 \pm 2.16$\,s, and the score is
$90.82 \pm 1.79$\% (mean $\pm$ standard deviation).
The mean number of action batches is also stable at $5.86 \pm 0.06$
per task.
Timing components are defined in Appendix~\ref{app:timing-components}.

\Needspace{29\baselineskip}
\subsection{Task-Level Score Agreement}
\label{app:task-repeatability}

\begin{table}[H]
\centering
\small
\setlength{\tabcolsep}{4pt}
\begin{tabular}{lrrr}
\toprule
I/O & $k$ & Pass@$k$ (\%) & Best-of-$k$ partial score (\%) \\
\midrule
Standard I/O & 1 & 87.20 & 90.82 \\
 & 2 & 88.00 & 91.62 \\
 & 3 & 88.00 & 91.62 \\
 & 4 & 88.00 & 91.62 \\
 & 5 & 88.00 & 91.62 \\
\midrule
Fast I/O & 1 & 86.80 & 90.42 \\
 & 2 & 88.80 & 92.42 \\
 & 3 & 89.20 & 92.82 \\
 & 4 & 89.60 & 93.22 \\
 & 5 & 90.00 & 93.62 \\
\bottomrule
\end{tabular}
\caption{\textbf{Task-level repeatability across five seeds of Astra at low effort.} Values average uniformly over all size-$k$ subsets of the five runs, then over the 50 tasks. Pass@$k$ measures whether any of the $k$ attempts succeeds; best-of-$k$ uses the highest partial score.}
\label{tab:repeatability}
\end{table}

Table~\ref{tab:repeatability} shows that additional attempts improve
performance only slightly. With standard I/O, 42 tasks receive full
credit in all five runs, two in four runs, and six in none.
Partial-credit scores are identical across all five runs on 48 of the
50 tasks. The low variation in mean score therefore reflects
consistent performance on individual tasks: almost every task receives
the same score on every attempt. With Fast I/O, 40 tasks receive full
credit in all five runs, four in four runs, one in one run, and five
in none. Partial-credit scores are identical on 45 tasks.

\paragraph{Pass@$k$ and partial credit.}
For a task with $c$ fully successful runs among the five, the probability
that a uniformly selected set of $k$ runs contains a success is
$1-\binom{5-c}{k}/\binom{5}{k}$. Pass@$k$ averages this quantity over
tasks. For partial credit, the table averages the maximum
task score within each size-$k$ set. Standard-I/O pass@1 is 87.2\%
and pass@5 is 88.0\%; its best-of-five partial score is 91.62\%.

\subsection{Variation in Action Sequences}

We next compare the number of action batches used for the same task
across seeds. For each task, we compute the standard deviation and
range across its five trajectories. With standard I/O, the median of
these task-level standard deviations is 0.55 batches, and the median
range is one batch. For Fast I/O, the corresponding values are 0.99
and two batches.

\paragraph{Two ways to complete the same spreadsheet task.}
For example, on the Calc task that cleans movie titles, all five
standard-I/O runs receive full credit while taking 7, 6, 5, 6, and
2 batches.
The seven-batch trajectory enters separate
\texttt{PROPER(TRIM(...))} formulas for successive rows and revisits
the paste-special dialog. The two-batch trajectory selects
\texttt{C2:C29}, enters one formula with Alt+Enter to fill the
selection, then saves. The two trajectories take 97.4 and 40.3\,s,
respectively. Here, the agent reduces the number of batches by
applying the formula to all rows at once.

\clearpage
\section{Representative Task Selection}
\label{app:selection}

We select a small set of tasks from each benchmark to reduce evaluation
time while preserving the scores and relative performance of agents.
This section gives the selection procedure from
Sec.~\ref{sec:representative-tasks} and
tests its score estimates, rankings, and speed--performance frontier
on held-out models.
Table~\ref{tab:representative-task-counts} lists the resulting task counts.

\begin{table}[H]
    \centering
    \begin{tabular}{lrr}
        \toprule
        Benchmark & Full set & Selected subset \\
        \midrule
        OSWorld~\citep{xie2024osworldbenchmarkingmultimodalagents} & 295 & 50 \\
        OSWorld2~\citep{yuan2026osworld20benchmarkingcomputer} & 63 & 52 \\
        CUA-World~\citep{aggarwal2026gymanythingturnsoftwareagent} & 143 & 26 \\
        MyPCBench~\citep{jang2026mypcbenchbenchmarkpersonallyintelligent} & 184 & 38 \\
        \bottomrule
    \end{tabular}
    \caption{Task counts for the representative subsets selected for \BenchmarkName{}.}
    \label{tab:representative-task-counts}
\end{table}

\subsection{Models and Tasks Used for Selection}

\paragraph{OSWorld.}
We use nine configurations evaluated on the same
295 tasks: Claude Sonnet 5 xhigh, Gemini 3.6 Flash, Gemini 3 Flash Preview,
GLM-5V Turbo, GPT-5.6 Luna xhigh, Kimi K3, Meta Muse Spark, MiniMax M3,
and Qwen3.5-9B-Thinking. Each task is represented by its nine partial
scores and nine exact-completion indicators. The internet-dependence
audit defining this task population is described in
Appendix~\ref{app:internet-audit}.

\paragraph{OSWorld2.}
Selection uses seven configurations:
MiniMax-M3, Claude Opus 4.7, Claude Sonnet 4.6 at medium and maximum
effort, GPT-5.5, GPT-5.6, and Qwen3.7. The resulting evaluation set
contains 52 tasks.

\paragraph{CUA-World.}
Selection uses seven configurations with scores on more
than 30 tasks each: Qwen3-VL-2B-Thinking, GPT-5.4 through Azure,
GPT-5.4, Claude Opus 4.7, Claude Sonnet 4.6, Gemini 3 Flash Preview,
and Kimi K2.5. Selection uses each model's distribution of observed
partial scores. Repeated runs remain separate observations. The
selected set contains 26 tasks.

\paragraph{MyPCBench.}
The evaluation uses 38 of the 184 canonical tasks, selected with the
unweighted energy method and 100 initializations. The task identities
for all four benchmarks are listed in
Appendix~\ref{app:task-membership}.

\subsection{Initialization, Optimization, and Task Count}

We seek a subset that represents the full benchmark's distribution of
score patterns. For
OSWorld and OSWorld2, let $z_i$ concatenate the partial
scores and exact-completion indicators for task $i$. Distances are
$d_{ij}=\lVert z_i-z_j\rVert_2$. For a subset $S$ of size $K$ from
$N$ tasks, the normalized energy objective is
\begin{equation}
 E(S)=\frac{1}{\bar d}\left[
 \frac{2}{KN}\sum_{i\in S}\sum_{j=1}^{N}d_{ij}
 -\frac{1}{K^2}\sum_{i,j\in S}d_{ij}
 -\frac{1}{N^2}\sum_{i,j=1}^{N}d_{ij}\right],
 \qquad
 \bar d=\frac{\sum_{i\ne j}d_{ij}}{N(N-1)}.
\end{equation}
We estimate a model's benchmark score by its unweighted mean score
on the selected tasks.

\paragraph{Balanced initializations.}
To cover a range of score profiles in each initial subset, we sort
tasks by the leading principal coordinate of their centered
score profiles and choose evenly spaced tasks from this ordering.
Specifically, for a uniformly sampled phase $\phi\in[0,1)$, the
initial positions are $\lfloor(\phi+t)N/K\rfloor$ for
$t=0,\ldots,K-1$. We use 100 initializations with seeds
$7000,\ldots,7099$, optimizing each distinct initial subset once.

\paragraph{One-swap optimization.}
At each iteration, we evaluate every exchange of a selected task with
an unselected task and apply the exchange that reduces energy the most.
We stop when no exchange improves the objective beyond numerical
tolerance and retain the lowest-energy subset across the initializations.

\Needspace{20\baselineskip}
\begin{agentlisting}{Representative-task selection and validation.\label{lst:selection}}
for each candidate task count K:
    for each model h:
        remove model h from the calibration models
        form task profiles from the remaining models
        optimize 100 balanced initial subsets by one-task swaps
        select the subset with the lowest calibration energy
        predict h's partial and exact scores by subset means
    compare predictions with each model's full-set means

choose the smallest K for which both rank correlations
are at least 0.95 at K-1, K, and K+1
refit the K-task subset using all models
\end{agentlisting}

To choose the number of tasks, we repeat the full selection procedure
with each model held out in turn
(Listing~\ref{lst:selection}). We construct task profiles and select
tasks using the remaining models, then compare the held-out model's
subset score with its full-set score. We choose the smallest $K$ for
which both partial-score and exact-completion rank correlations are
at least 0.95 at $K-1$, $K$, and $K+1$. Checking the neighboring task
counts tests the stability of the ranking criterion. Each
task count is optimized independently. After choosing $K$, we select the final evaluation set
using all models.

\paragraph{Selection with missing scores.}
Some CUA-World models have scores on only part of the benchmark. We
therefore compare each model's distribution of observed partial scores
on selected tasks with its distribution over all observed tasks,
using one-dimensional energy distance. We first ensure that the subset
contains an observation for every model used for selection, then
minimize the mean of these distances. The task-count rule uses partial-score rank correlation
at $K-1,K,K+1$.
At $K=25,26,27$, the seven-model held-out correlations are
0.964, 1.000, and 1.000, respectively; the corresponding score errors
are 7.43, 5.12, and 2.10 percentage points. These errors compare
unweighted means of observed run scores.

\clearpage
\begin{figure}[!t]
  \centering
  \includegraphics[width=\linewidth]{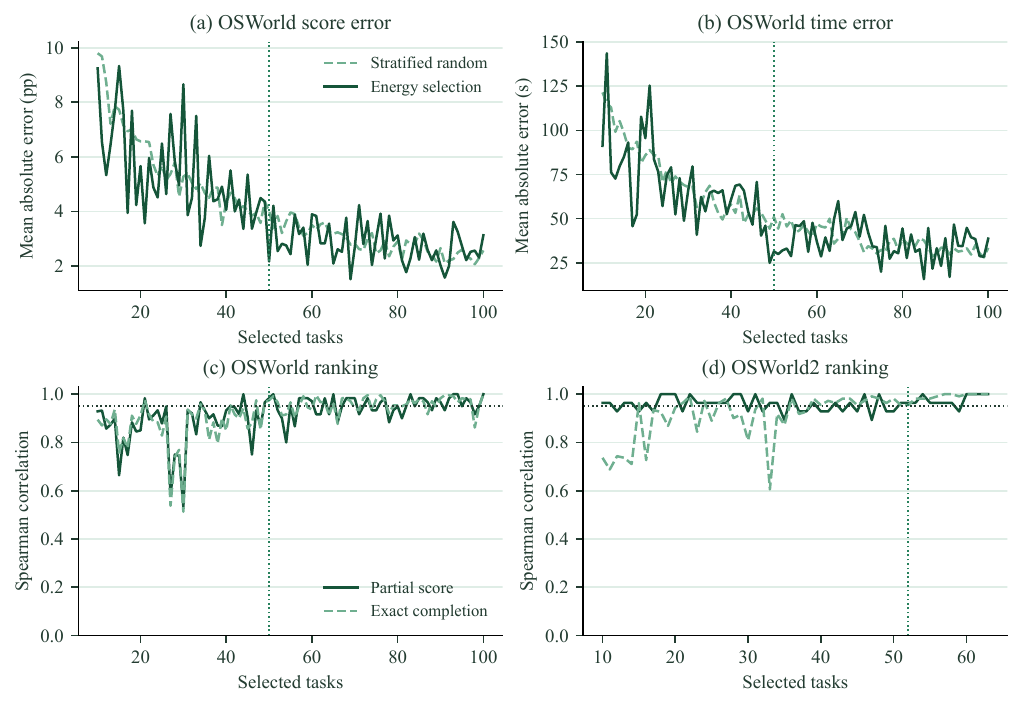}
  \caption{\textbf{Score error and rank correlation on held-out models.}
  OSWorld uses nine leave-one-agent-out folds; OSWorld2 uses seven.
  The randomized baseline averages 300 difficulty-stratified selections
  per task count. Dotted vertical lines mark 50 and 52 tasks; horizontal
  lines mark rank correlation 0.95. Each task count is optimized
  independently.}
  \label{fig:selection-validation}
\end{figure}

\subsection{Held-Out Score and Ranking Accuracy}

Figure~\ref{fig:selection-validation} compares energy-based selection
with a randomized baseline that samples tasks across
difficulty levels. We measure error as the mean absolute difference
between each held-out model's subset estimate and its full-set result.
At 50 OSWorld tasks, energy-based selection reduces
partial-score error from 4.02 to 2.16 percentage points. It also
reduces mean-task-time error from 51.0 to 31.4\,s, even though
selection uses only scores. Thus, selecting tasks with representative
score patterns also improves estimates of execution time.

\begin{table}[H]
\centering
\small
\setlength{\tabcolsep}{4pt}
\begin{tabular}{lrrrrr}
\toprule
Benchmark & $K$ & Partial $\rho$ & Exact $\rho$ & Partial MAE & Exact MAE \\
\midrule
OSWorld & 49 & 0.967 & 0.967 & 4.36 & 4.57 \\
 & 50 & 0.983 & 0.975 & 2.16 & 2.17 \\
 & 51 & 1.000 & 0.992 & 4.21 & 3.95 \\
\midrule
OSWorld2 & 51 & 0.964 & 0.955 & 1.70 & 2.28 \\
 & 52 & 0.964 & 0.955 & 1.33 & 2.03 \\
 & 53 & 0.964 & 0.982 & 0.74 & 1.04 \\
\bottomrule
\end{tabular}
\caption{\textbf{Held-out-model validation around the selected task count.} MAE is in percentage points. Each model is excluded from task selection in its fold. Both correlations exceed 0.95 at the selected task count and its two neighbors.}
\label{tab:selection-validation}
\end{table}

Table~\ref{tab:selection-validation} shows that the selected task sets
closely preserve model rankings on both benchmarks. At 50 tasks, OSWorld
achieves partial-score and exact-completion rank correlations of
0.983 and 0.975; at 52 tasks, OSWorld2 achieves 0.964 and 0.955.
Both correlations remain above 0.95 at the neighboring task counts.

\clearpage
\begin{figure}[!t]
  \centering
  \includegraphics[width=\linewidth]{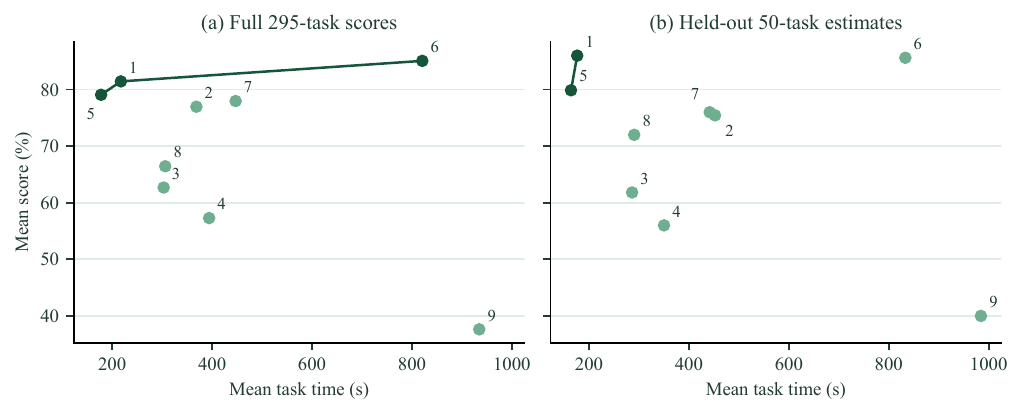}
  \caption{\textbf{Speed--performance frontiers from full-set scores and
  held-out subset estimates.} Dark points and connecting lines mark
  nondominated configurations. Model IDs are: 1, Sonnet 5; 2, Gemini 3.6
  Flash; 3, Gemini 3 Flash Preview; 4, GLM-5V Turbo; 5, Luna;
  6, Kimi K3; 7, Muse Spark; 8, MiniMax M3; 9, Qwen3.5-9B.
  Each point in (b) uses a 50-task subset selected with that model held out.}
  \label{fig:selection-frontier}
\end{figure}

\subsection{Preserving the Speed--Performance Frontier}

Figure~\ref{fig:selection-frontier} compares the full-set frontier with
the frontier estimated from 50-task subsets for held-out models. The
estimated frontier recovers two of the three full-set frontier
configurations and adds no others, giving 100\% precision and 66.7\%
recall. Pairwise dominance agrees for 91.7\% of model pairs. This analysis
uses the models' scores and times; selection itself uses only scores.

\paragraph{Evaluation savings.}
Selecting 50 of the 295 OSWorld tasks reduces the number of tasks by
$5.9\times$. Across the nine models, these tasks account
for 15.1\% of the total recorded task time, a $6.61\times$ reduction.
We compute this saving by summing trajectory times, excluding setup
and verification.

\clearpage
\begin{table}[!t]
\centering
\small
\begin{tabular}{lrrr}
\toprule
Method & Partial MAE & Exact MAE & Mean MAE \\
\midrule
Uniform random & 5.88 & 6.01 & 5.95 \\
Difficulty-stratified random & 5.05 & 5.11 & 5.08 \\
IRT-inspired selection & 7.94 & 8.21 & 8.08 \\
Energy-based selection & \textbf{4.50} & \textbf{4.34} & \textbf{4.42} \\
\bottomrule
\end{tabular}
\caption{\textbf{Task-selection methods on OSWorld with 32 tasks.}
Errors are averaged across nine leave-one-agent-out folds and reported
in percentage points. Mean MAE averages partial-score and
exact-completion MAE. Random baselines average 300 selections per fold.}
\label{tab:selection-baselines}
\end{table}

\subsection{Comparing Task-Selection Methods}
\label{app:selection-baselines}

Table~\ref{tab:selection-baselines} compares task-selection methods at a
budget of 32 OSWorld tasks, using the same nine leave-one-agent-out
folds. Energy-based selection achieves a mean error of 4.42 percentage
points, compared to 5.08 for difficulty-stratified random selection and
8.08 for the IRT-inspired method.

We use energy-based selection for its simple scoring rule: each agent's
benchmark score is the average of its scores on the selected tasks.
The IRT-inspired method fits a one-parameter ability model
using the eight agents available in each fold and predicts full-benchmark
performance from the selected-task scores.

\subsection{Selected Task Identities}
\label{app:task-membership}

The following lists give the final task sets used in our evaluations.
OSWorld IDs use the original task UUIDs. OSWorld2 uses its three-digit identifiers;
MyPCBench IDs combine the task family and suffix; CUA-World IDs combine
the environment and task name.

\Needspace{9\baselineskip}
\subsubsection{OSWorld}
\label{app:tasks-osworld-energy50-representative}
\begingroup\footnotesize
\setlength{\tabcolsep}{4pt}
\begin{longtable}{p{0.29\linewidth}p{0.65\linewidth}}
\caption{Selected OSWorld task identities.}\label{tab:tasks-osworld-energy50-representative}\\
\toprule
Application & Task UUID \\
\midrule
\endfirsthead
\toprule
Application & Task UUID \\
\midrule
\endhead
\bottomrule
\endlastfoot
chrome & \texttt{06fe7178-4491-4589-810f-2e2bc9502122} \\
chrome & \texttt{2ae9ba84-3a0d-4d4c-8338-3a1478dc5fe3} \\
chrome & \texttt{3720f614-37fd-4d04-8a6b-76f54f8c222d} \\
chrome & \texttt{44ee5668-ecd5-4366-a6ce-c1c9b8d4e938} \\
chrome & \texttt{af630914-714e-4a24-a7bb-f9af687d3b91} \\
gimp & \texttt{045bf3ff-9077-4b86-b483-a1040a949cff} \\
gimp & \texttt{554785e9-4523-4e7a-b8e1-8016f565f56a} \\
gimp & \texttt{734d6579-c07d-47a8-9ae2-13339795476b} \\
gimp & \texttt{a746add2-cab0-4740-ac36-c3769d9bfb46} \\
libreoffice\_\allowbreak calc & \texttt{0326d92d-d218-48a8-9ca1-981cd6d064c7} \\
libreoffice\_\allowbreak calc & \texttt{1de60575-bb6e-4c3d-9e6a-2fa699f9f197} \\
libreoffice\_\allowbreak calc & \texttt{1e8df695-bd1b-45b3-b557-e7d599cf7597} \\
libreoffice\_\allowbreak calc & \texttt{3a7c8185-25c1-4941-bd7b-96e823c9f21f} \\
libreoffice\_\allowbreak calc & \texttt{4de54231-e4b5-49e3-b2ba-61a0bec721c0} \\
libreoffice\_\allowbreak calc & \texttt{535364ea-05bd-46ea-9937-9f55c68507e8} \\
libreoffice\_\allowbreak calc & \texttt{7efeb4b1-3d19-4762-b163-63328d66303b} \\
libreoffice\_\allowbreak calc & \texttt{8b1ce5f2-59d2-4dcc-b0b0-666a714b9a14} \\
libreoffice\_\allowbreak calc & \texttt{a9f325aa-8c05-4e4f-8341-9e4358565f4f} \\
libreoffice\_\allowbreak calc & \texttt{abed40dc-063f-4598-8ba5-9fe749c0615d} \\
libreoffice\_\allowbreak impress & \texttt{39be0d19-634d-4475-8768-09c130f5425d} \\
libreoffice\_\allowbreak impress & \texttt{455d3c66-7dc6-4537-a39a-36d3e9119df7} \\
libreoffice\_\allowbreak impress & \texttt{986fc832-6af2-417c-8845-9272b3a1528b} \\
libreoffice\_\allowbreak impress & \texttt{a097acff-6266-4291-9fbd-137af7ecd439} \\
libreoffice\_\allowbreak impress & \texttt{a434992a-89df-4577-925c-0c58b747f0f4} \\
libreoffice\_\allowbreak impress & \texttt{af2d657a-e6b3-4c6a-9f67-9e3ed015974c} \\
libreoffice\_\allowbreak impress & \texttt{b8adbc24-cef2-4b15-99d5-ecbe7ff445eb} \\
libreoffice\_\allowbreak writer & \texttt{0810415c-bde4-4443-9047-d5f70165a697} \\
libreoffice\_\allowbreak writer & \texttt{0e47de2a-32e0-456c-a366-8c607ef7a9d2} \\
libreoffice\_\allowbreak writer & \texttt{72b810ef-4156-4d09-8f08-a0cf57e7cefe} \\
libreoffice\_\allowbreak writer & \texttt{adf5e2c3-64c7-4644-b7b6-d2f0167927e7} \\
libreoffice\_\allowbreak writer & \texttt{ecc2413d-8a48-416e-a3a2-d30106ca36cb} \\
multi\_\allowbreak apps & \texttt{42f4d1c7-4521-4161-b646-0a8934e36081} \\
multi\_\allowbreak apps & \texttt{48d05431-6cd5-4e76-82eb-12b60d823f7d} \\
multi\_\allowbreak apps & \texttt{6d72aad6-187a-4392-a4c4-ed87269c51cf} \\
multi\_\allowbreak apps & \texttt{81c425f5-78f3-4771-afd6-3d2973825947} \\
multi\_\allowbreak apps & \texttt{d9b7c649-c975-4f53-88f5-940b29c47247} \\
multi\_\allowbreak apps & \texttt{e135df7c-7687-4ac0-a5f0-76b74438b53e} \\
multi\_\allowbreak apps & \texttt{f5c13cdd-205c-4719-a562-348ae5cd1d91} \\
os & \texttt{b6781586-6346-41cd-935a-a6b1487918fc} \\
thunderbird & \texttt{7b1e1ff9-bb85-49be-b01d-d6424be18cd0} \\
thunderbird & \texttt{9b7bc335-06b5-4cd3-9119-1a649c478509} \\
thunderbird & \texttt{a1af9f1c-50d5-4bc3-a51e-4d9b425ff638} \\
thunderbird & \texttt{dfac9ee8-9bc4-4cdc-b465-4a4bfcd2f397} \\
thunderbird & \texttt{f201fbc3-44e6-46fc-bcaa-432f9815454c} \\
vlc & \texttt{cb130f0d-d36f-4302-9838-b3baf46139b6} \\
vlc & \texttt{d06f0d4d-2cd5-4ede-8de9-598629438c6e} \\
vlc & \texttt{fba2c100-79e8-42df-ae74-b592418d54f4} \\
vs\_\allowbreak code & \texttt{0ed39f63-6049-43d4-ba4d-5fa2fe04a951} \\
vs\_\allowbreak code & \texttt{5e2d93d8-8ad0-4435-b150-1692aacaa994} \\
vs\_\allowbreak code & \texttt{ec71221e-ac43-46f9-89b8-ee7d80f7e1c5} \\
\end{longtable}
\endgroup

\Needspace{9\baselineskip}
\subsubsection{OSWorld2}
\label{app:tasks-osworld2-k52}
The selected task IDs are \texttt{001}, \texttt{002}, \texttt{010}, \texttt{012}, \texttt{013}, \texttt{015}, \texttt{018}, \texttt{020}, \texttt{021}, \texttt{022}, \texttt{028}, \texttt{029}, \texttt{030}, \texttt{033}, \texttt{038}, \texttt{040}, \texttt{042}, \texttt{043}, \texttt{044}, \texttt{046}, \texttt{047}, \texttt{049}, \texttt{051}, \texttt{053}, \texttt{054}, \texttt{057}, \texttt{058}, \texttt{059}, \texttt{061}, \texttt{063}, \texttt{065}, \texttt{066}, \texttt{068}, \texttt{070}, \texttt{071}, \texttt{072}, \texttt{076}, \texttt{080}, \texttt{085}, \texttt{086}, \texttt{088}, \texttt{091}, \texttt{094}, \texttt{096}, \texttt{100}, \texttt{101}, \texttt{102}, \texttt{103}, \texttt{104}, \texttt{106}, \texttt{107}, \texttt{108}.

\Needspace{9\baselineskip}
\subsubsection{MyPCBench}
\label{app:tasks-mypcbench-energy38}
\begin{center}\small
\begin{tabular}{ll}
\toprule
Task family & Selected IDs \\
\midrule
aggregation & f002, f005, f029, f033 \\
contradiction & f015, f023 \\
counterfactual & f001, f004, f008 \\
cua\_only & f004, f006, f018, f021, f023 \\
hard\_app & f011, f019, f022 \\
long\_horizon & f028, f043, f045, f047, f050, f060, f074 \\
preference\_inference & f004, f014, f021, f023, f025 \\
retrieval & f005, f029 \\
situated\_action & f001, f003, f006, f017, f026, f035, f039 \\
\bottomrule
\end{tabular}\end{center}

\Needspace{9\baselineskip}
\subsubsection{CUA-World}
\label{app:tasks-cua-world-long-k26}
\begingroup\footnotesize
\setlength{\tabcolsep}{4pt}
\begin{longtable}{p{0.29\linewidth}p{0.65\linewidth}}
\caption{Selected CUA-World task identities.}\label{tab:tasks-cua-world-long-k26}\\
\toprule
Environment & Task name \\
\midrule
\endfirsthead
\toprule
Environment & Task name \\
\midrule
\endhead
\bottomrule
\endlastfoot
ardour\_\allowbreak env & \texttt{broadcast\_\allowbreak podcast\_\allowbreak stem\_\allowbreak delivery} \\
dhis2\_\allowbreak env & \texttt{rmncah\_\allowbreak scorecard\_\allowbreak dashboard} \\
docker\_\allowbreak desktop\_\allowbreak env & \texttt{diagnose\_\allowbreak broken\_\allowbreak microservices\_\allowbreak stack} \\
gpredict\_\allowbreak env & \texttt{poes\_\allowbreak downlink\_\allowbreak schedule\_\allowbreak setup} \\
gvsig\_\allowbreak desktop\_\allowbreak env & \texttt{vulnerability\_\allowbreak map\_\allowbreak remote\_\allowbreak communities} \\
jstock\_\allowbreak env & \texttt{quarterly\_\allowbreak portfolio\_\allowbreak rebalance} \\
librehealth\_\allowbreak ehr\_\allowbreak env & \texttt{implement\_\allowbreak lab\_\allowbreak workflow\_\allowbreak and\_\allowbreak process\_\allowbreak patient} \\
moodle\_\allowbreak env & \texttt{configure\_\allowbreak tiered\_\allowbreak assessment\_\allowbreak pathway} \\
nextgen\_\allowbreak connect\_\allowbreak integration\_\allowbreak engine\_\allowbreak env & \texttt{adt\_\allowbreak census\_\allowbreak lab\_\allowbreak validation\_\allowbreak pipeline} \\
nosh\_\allowbreak env & \texttt{care\_\allowbreak quality\_\allowbreak remediation} \\
odoo\_\allowbreak inventory\_\allowbreak env & \texttt{pharma\_\allowbreak lot\_\allowbreak recall\_\allowbreak quarantine} \\
openclinic\_\allowbreak ga\_\allowbreak env & \texttt{insured\_\allowbreak consultation\_\allowbreak billing} \\
oracle\_\allowbreak database\_\allowbreak env & \texttt{claims\_\allowbreak pipeline\_\allowbreak reconciliation} \\
project\_\allowbreak libre\_\allowbreak env & \texttt{schedule\_\allowbreak recovery\_\allowbreak rebaseline} \\
pymol\_\allowbreak env & \texttt{kinase\_\allowbreak selectivity\_\allowbreak comparison} \\
redmine\_\allowbreak env & \texttt{q1\_\allowbreak milestone\_\allowbreak reconciliation} \\
rocket\_\allowbreak chat\_\allowbreak env & \texttt{compliance\_\allowbreak audit\_\allowbreak remediation} \\
snap\_\allowbreak env & \texttt{multicriteria\_\allowbreak suitability\_\allowbreak mapping} \\
splunk\_\allowbreak env & \texttt{threat\_\allowbreak intel\_\allowbreak enrichment\_\allowbreak pipeline} \\
sumo\_\allowbreak env & \texttt{optimize\_\allowbreak network\_\allowbreak signal\_\allowbreak timing} \\
thunderbird\_\allowbreak env & \texttt{litigation\_\allowbreak email\_\allowbreak triage} \\
wireshark\_\allowbreak env & \texttt{web\_\allowbreak app\_\allowbreak breach\_\allowbreak investigation} \\
wondershare\_\allowbreak edrawmax\_\allowbreak env & \texttt{healthcare\_\allowbreak it\_\allowbreak architecture\_\allowbreak review} \\
woo\_\allowbreak commerce\_\allowbreak env & \texttt{launch\_\allowbreak coffee\_\allowbreak product\_\allowbreak line} \\
wordpress\_\allowbreak env & \texttt{launch\_\allowbreak woocommerce\_\allowbreak coffee\_\allowbreak roastery} \\
wps\_\allowbreak presentation\_\allowbreak env & \texttt{rebrand\_\allowbreak restructure\_\allowbreak pitch\_\allowbreak deck} \\
\end{longtable}
\endgroup

\FloatBarrier
\section{Internet-Dependence Audit}
\label{app:internet-audit}

Live public services can change their content, become unavailable, or
restrict access between evaluations. We therefore audit whether each
task requires the agent to use such services. Tasks that use a local
application or a benchmark-controlled website remain eligible, as do
tasks that only require downloading fixed files during setup.

\subsection{Inclusion Rule}

We first use a coding agent to classify the 369 OSWorld tasks as
local/offline, benign internet, or anti-bot/access-sensitive. Three
human reviewers then independently assess whether each task requires
interaction with a live public service. We include a task only when
all three reviewers agree to include it. This gives 295 tasks.

\begin{table}[H]
  \centering\small
  \begin{tabular}{lrr}
    \toprule
    Reviewer or outcome & Included & Excluded \\
    \midrule
    Reviewer 1 & 297 & 72 \\
    Reviewer 2 & 302 & 67 \\
    Reviewer 3 & 295 & 74 \\
    \midrule
    Unanimous decision & 295 & 67 \\
    Disagreement & \multicolumn{2}{c}{7} \\
    \bottomrule
  \end{tabular}
  \caption{\textbf{Independent inclusion decisions for 369 OSWorld tasks.}
  The seven disagreements are excluded by the unanimous-inclusion rule.}
  \label{tab:internet-votes}
\end{table}

\subsection{Reviewer Agreement}

Table~\ref{tab:internet-votes} shows that 362 of 369 tasks receive a
unanimous decision. Pairwise agreement is 98.64\%, 99.46\%, and 98.10\%
for reviewer pairs 1--2, 1--3, and 2--3, respectively. Their Cohen's
$\kappa$ values are 0.956, 0.983, and 0.939; Fleiss' $\kappa$ across
all three reviewers is 0.959.

\paragraph{Applying the criterion across benchmarks.}
We apply the same criterion to the other benchmarks. OSWorld2 provides
benchmark-controlled web applications, and MyPCBench provides local,
pre-authenticated services, so these tasks remain eligible. For
CUA-World, we exclude tasks requiring internet access during agent
interaction while retaining those that only use the network for setup.

\FloatBarrier
\Needspace{14\baselineskip}
\section{Validating Actions with CUA-AutoDebug}
\label{app:autoharness}

A task can fail because the model chooses an incorrect action or because
the infrastructure executes a correct action incorrectly. We developed
CUA-AutoDebug to distinguish these errors. A controlled desktop
application records the keyboard and mouse inputs it receives, including
modifier keys and typed text. We also check the resulting application
state: a click should reach the intended target, a drag should move the
object, and typed text should match the requested string. For shortcuts
handled by the operating system, we record input events before they reach
the application.

\subsection{Separating Model, Harness, and Runtime Errors}

We first submit predefined actions directly to the VM runtime.
We then test agent harnesses by giving the model a short instruction,
such as clicking a colored square, and comparing its response with the
executed action and received input. An incorrect model action is a model
error; incorrect translation of a correct action is a harness error.

\subsection{Input Failures Detected by the Runtime Tests}

\begin{table}[H]
  \centering\small
  \setlength{\tabcolsep}{4pt}
  \begin{tabular}{p{0.23\linewidth}p{0.34\linewidth}p{0.32\linewidth}}
    \toprule
    Requested input & Received input & Cause \\
    \midrule
    Type \texttt{<} & \texttt{>} & Incorrect X11 key/modifier mapping \\
    Type accented text & Accented characters omitted & Characters absent from the key map \\
    Type a literal shell variable & Variable expanded into a path & Text interpreted by the command shell \\
    Keypad Enter or Menu & No corresponding key event & Missing key-name translation \\
    Toggle Caps Lock, then type & Lowercase text & Caps Lock press silently omitted \\
    \bottomrule
  \end{tabular}
  \caption{\textbf{Input errors detected by CUA-AutoDebug when executing
  actions through SSH and PyAutoGUI.} The tests compare requested input
  with the keyboard and mouse events received.}
  \label{tab:autoharness-bugs}
\end{table}

Table~\ref{tab:autoharness-bugs} shows examples of input errors detected
by these tests, such as typing \texttt{>} when \texttt{<} was requested or
silently dropping accented characters. We test a catalog of 100 cases
covering clicks, drags, scrolls, key presses, key combinations, text entry,
and sequences of these actions. Six cases are unsupported by the tested
action interface. We repeat each of the remaining 94 cases five times;
83 pass and 11 fail when executed through SSH and PyAutoGUI. The failures
include one shell-expansion case, one shifted-symbol case, five Unicode
cases, and four named-key cases.

\paragraph{The corrected runtime passes all 470 tests.}
We correct these errors in CUA-Speedrun's QEMU runtime by preserving
literal text when sending commands, explicitly mapping key names and
modifiers, and supporting characters absent from the default keyboard
mapping. After these corrections, all 94 supported cases pass in all
five repetitions.

\subsection{Failures in Agent Harnesses}

\begin{table}[H]
  \centering\small
  \setlength{\tabcolsep}{4pt}
  \begin{tabular}{p{0.20\linewidth}p{0.35\linewidth}p{0.34\linewidth}}
    \toprule
    Harness & Model response & Executed behavior \\
    \midrule
    Gemini & Scroll down by five wheel clicks & Scroll up by 600 ticks \\
    Gemini & Press F5 or Page Down & Type the key name as text \\
    Qwen3.5 & Middle-click the target & No action \\
    Qwen3.5 & Ctrl-click the target & Release Ctrl before clicking \\
    Qwen3.5 & Move, then scroll or drag & Execute only the first tool call \\
    \bottomrule
  \end{tabular}
  \caption{\textbf{Harness errors can change a correct model action.}
  Examples detected by comparing model responses with the actions
  executed and the input events received.}
  \label{tab:autoharness-agent-bugs}
\end{table}

Table~\ref{tab:autoharness-agent-bugs} shows that harness errors can
prevent a correct model response from reaching the application. For
example, the Gemini harness reversed scroll direction and interpreted a
fallback scroll distance in pixels as a number of wheel ticks. All six
scroll-direction test repetitions failed before correction and passed
afterward. The Qwen3.5 harness released the modifier key before a click
and silently discarded tool calls after the first. A model that described
a triple-click but emitted a double-click produced a model error.

\FloatBarrier
\section{Experimental Details}
\label{app:experimental-specification}

Appendix~\ref{app:configurations}
lists the model, harness, and reasoning setting for each configuration.

\subsection{Task Definition}
\label{app:task-definition}

We define a computer-use task as $x_i=(E_i,s_i^0,p_i,V_i)$, where $E_i$ is
an interactive environment with initial state $s_i^0$, $p_i$ is a
natural-language instruction, and $V_i$ is a verification function. An
agent $\pi$ receives $p_i$ and a sequence of observations and produces
computer actions until it terminates or reaches the task limit. This
interaction produces a trajectory $\tau_i$ and final state $s_i^T$, with
score $r_i=V_i(\tau_i,s_i^T)$. A benchmark $B=\{x_i\}_{i=1}^{N}$ is a
collection of such tasks.

\paragraph{Agent implementations.}
We connect each model's public reference agent or native computer-use API
to the common desktop interface.

\subsection{Observations, Actions, and Interaction History}

The environment provides $1920\times1080$ screenshots of the desktop.
Mouse actions use pixel coordinates in these images.
Harnesses that resize screenshots or use normalized coordinates convert
the model's coordinates to desktop pixels before executing an action.

\begin{table}[H]
  \centering\small
  \setlength{\tabcolsep}{4pt}
  \begin{tabular}{p{0.22\linewidth}p{0.67\linewidth}}
    \toprule
    Action family & Operations \\
    \midrule
    Mouse clicks & Left, right, middle, double, and triple click at a coordinate \\
    Pointer motion & Move, drag along coordinates, hold or release a mouse button \\
    Scrolling & Signed vertical scroll at the current pointer position \\
    Keyboard & Type text, press a key or chord, hold and release modifiers \\
    Waiting & Wait for an agent-specified duration \\
    Observation & Request a screenshot without changing the desktop \\
    Completion & Stop interaction and invoke the benchmark verifier \\
    \bottomrule
  \end{tabular}
  \caption{\textbf{Desktop actions supported by the infrastructure.} A batch
  contains one or more actions executed in order. The model-facing
  tool schema can differ between harnesses.}
  \label{tab:action-vocabulary}
\end{table}

Table~\ref{tab:action-vocabulary} lists the actions available to agents.
Direct-API agents translate the model's computer-use responses into these
actions and return screenshots in the provider's tool-result format.
Codex uses the same actions from a separate agent sandbox. We instruct
Codex to interact with the task desktop only through the supplied
interface: its local shell and files belong to the agent sandbox
(Listing~\ref{lst:codex-isolation}).

\Needspace{12\baselineskip}
\begin{agentlisting}{Core desktop-isolation instruction in the Codex prompt.\label{lst:codex-isolation}}
You are operating a remote computer in a separate VM.
Your only interface to that computer is the HTTP proxy at $CS_COMPUTER_URL.
Authenticate every request with the header X-Gateway-Token: $CS_COMPUTER_TOKEN.
Your local shell and files are in the agent sandbox, not the task VM.
Do not bypass the proxy to access the task computer.
\end{agentlisting}

\paragraph{Interaction history.}
Each harness determines which screenshots and previous responses remain
in the model's context. Kimi's harness retains the current
screenshot and the two most recent completed-turn screenshots.
Earlier turns are compacted into textual history; recent responses
retain their reasoning fields and tool calls.
Codex maintains its own interaction history and explicitly retrieves
screenshots.

\paragraph{Reasoning settings.}
We vary reasoning effort within each model using the settings exposed by
its provider, such as low and high.

\subsection{Task Limits and Execution}

Each evaluation fixes the task set, agent implementation, reasoning
setting, environment resources, and number of parallel tasks. Every task
starts from its benchmark-specified initial state. We measure task time
from when the instruction is given to the agent until it terminates or
reaches the task limit, excluding setup, initialization, and verification
(Sec.~\ref{sec:standardized-cua-eval}).

\paragraph{Time and action limits.}
The OSWorld and OSWorld2 configurations in Appendix~\ref{app:configurations}
have an environment deadline of 39,600\,s per task, while the MyPCBench
Astra sweep uses 7,200\,s. These deadlines apply separately from the harness's action or
model-turn limit and the timeout on an individual model request.

We allow 500 action batches per task in the Astra batching comparison,
five-seed experiments, and CUA-World sweep, and 100 in the MyPCBench
sweep. A batch can contain several actions before the next observation;
the single-action Astra ablation requires a screenshot after
every action (Appendix~\ref{app:batching}). In the Fast I/O comparison,
we keep the model and reasoning setting fixed and change the action and
screenshot implementation (Appendix~\ref{app:fast-io-profile}).

\subsection{Benchmark Verification}

We use each benchmark's verification procedure. OSWorld and OSWorld2
score task completion with task-specific programmatic checks. CUA-World
uses Gemini 3 Flash Preview to assess the recorded trajectory against
its visual checklist. MyPCBench uses its visual trajectory judge and
averages the rubric scores with equal weight; a task receives full
credit only when every rubric passes. We also retain each benchmark's
task instructions and context, including MyPCBench's persona and
pre-authenticated local applications, alongside the agent's system prompt.

\subsection{Model Cost}

We compute model cost by summing the cost of each task's model
requests. Each request's cost comes from the provider's recorded charge
or its recorded token usage at the applicable API prices.
Generated tokens follow the accounting in
Appendix~\ref{app:token-accounting}.

Model cost excludes environment hosting, model-server
rental, setup, and separate verification calls.

\FloatBarrier
\section{Additional Agent Analyses}
\label{app:additional-analyses}

We examine how action batching, agent harnesses, and reasoning effort
affect task time through comparisons within the same model. Action
batches, individual actions, and generated tokens are defined in
Appendix~\ref{app:measurements}.

\subsection{Controlled Batching Comparisons}
\label{app:batching}

\begin{table}[H]
\centering
\footnotesize
\setlength{\tabcolsep}{4pt}
\begin{tabular}{ll l rrrrr}
\toprule
Model & Effort & Interaction rule & Score & Time & Batches & Actions & Tokens \\
\midrule
Astra & xhigh & Batched & 91.62 & 126.82 & 7.14 & 19.90 & 2,060 \\
Astra & xhigh & One action + image & 91.62 & 182.73 & 16.50 & 16.50 & 2,559 \\
\midrule
Kimi K3 & low & Single tool & 71.62 & 176.61 & 8.12 & 55.10 & 3,171 \\
Kimi K3 & low & Batched tools & 73.62 & 149.50 & 5.74 & 30.00 & 1,865 \\
\midrule
Kimi K3 & high & Single tool & 81.62 & 305.07 & 8.92 & 113.60 & 7,567 \\
Kimi K3 & high & Batched tools & 79.62 & 358.13 & 8.76 & 60.20 & 7,125 \\
\midrule
Kimi K3 & max & Single tool & 85.62 & 606.27 & 12.18 & 61.24 & 14,359 \\
Kimi K3 & max & Batched tools & 89.62 & 443.74 & 11.70 & 53.58 & 10,435 \\
\bottomrule
\end{tabular}
\caption{\textbf{Batching comparisons within the same model and reasoning effort on OSWorld.} Score is a percentage; time is in seconds. All counts and times are means over the same 50 tasks. Tokens include reasoning and other output.}
\label{tab:batching-ablation}
\end{table}

\paragraph{Batching makes Astra faster at the same score.}
Table~\ref{tab:batching-ablation} compares Astra xhigh with its standard
batched actions and a restriction to one action between screenshots.
Both settings score 91.62\%, but the single-action setting takes
182.7\,s per task compared with 126.8\,s for batching. It executes fewer
individual actions (16.50 versus 19.90), yet requires more action batches
(16.50 versus 7.14) and generates more tokens (2,559 versus 2,060).
We keep the model, reasoning effort, 50-task set, and configured limits
fixed. To enforce the single-action restriction, we reject multi-action
requests and prevent further actions until the agent retrieves a new
screenshot, including actions issued in separate back-to-back requests.

\paragraph{Kimi benefits from batched tools at some reasoning settings.}
Table~\ref{tab:batching-ablation} also compares Kimi at matched low,
high, and maximum effort. Kimi's single-tool setting disables parallel
tool calls and allows
multiple GUI actions within one code block. At maximum effort, batched
tools reduce mean task time from 606.3 to 443.7\,s and improve the score
from 85.62\% to 89.62\%. At high effort, batching increases time
from 305.1 to 358.1\,s while reducing the score from 81.62\% to 79.62\%.

\FloatBarrier
\clearpage
\begin{figure}[!t]
  \centering
  \includegraphics[width=\linewidth]{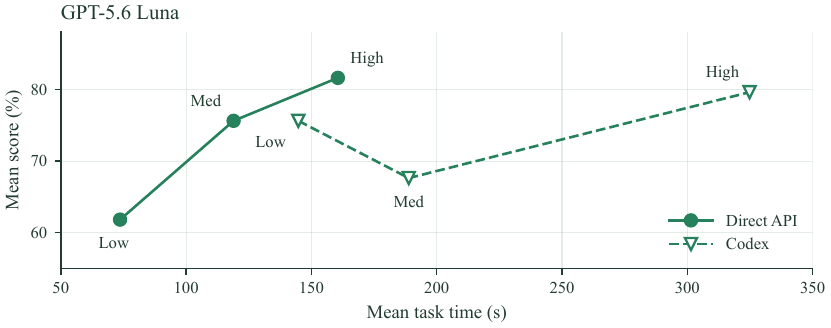}
  \caption{\textbf{Harness performance depends on reasoning effort.}
  GPT-5.6 Luna at low, medium, and high reasoning effort, using the
  direct-API agent or Codex on OSWorld. Each curve connects reasoning
  settings within one harness. All six evaluations use the same 50-task set.}
  \label{fig:results-harness-full}
\end{figure}

\subsection{Harness Comparisons Across Reasoning Efforts}
\label{app:harness-efforts}

Figure~\ref{fig:results-harness-full} compares GPT-5.6 Luna with Codex
and the direct-API agent at low, medium, and high effort. At low effort,
Codex improves the score from 61.8\% to 75.6\% but increases mean task
time from 74 to 145\,s. At medium and high effort, the direct-API agent
is both more accurate and faster: it scores 75.6\% in 119\,s compared
with 67.6\% in 189\,s at medium, and 81.6\% in 161\,s compared with
79.6\% in 325\,s at high.

\FloatBarrier
\Needspace{36\baselineskip}
\subsection{Comparing Reasoning Efforts on the Same Successful Tasks}
\label{app:common-success}

\begin{table}[H]
\centering
\footnotesize
\setlength{\tabcolsep}{4pt}
\begin{tabular}{lrlrrrrr}
\toprule
Model & Tasks & Effort & Time & Batches & Actions & Tokens & Tok/s \\
\midrule
Gemini 3.8 Flash & 5 & low & 1291.0 & 116.40 & 171.60 & 29,596 & 47.57 \\
 &  & medium & 1363.9 & 112.60 & 158.40 & 34,446 & 43.10 \\
 &  & high & 3024.5 & 190.60 & 246.60 & 73,578 & 42.18 \\
\midrule
GPT-6 Astra & 13 & low & 791.8 & 47.08 & 146.69 & 7,878 & 15.18 \\
 &  & medium & 752.6 & 47.15 & 138.85 & 8,376 & 17.07 \\
 &  & high & 817.3 & 53.15 & 171.23 & 11,001 & 21.31 \\
 &  & xhigh & 1131.6 & 55.38 & 223.46 & 14,359 & 18.04 \\
\midrule
Muse Spark 1.3 & 2 & minimal & 833.1 & 82.00 & 195.00 & 7,521 & 15.04 \\
 &  & low & 1183.3 & 99.50 & 291.00 & 12,997 & 16.79 \\
 &  & medium & 1008.8 & 78.50 & 235.00 & 14,621 & 21.55 \\
 &  & high & 907.3 & 68.50 & 246.00 & 14,095 & 22.87 \\
 &  & xhigh & 1546.2 & 109.50 & 400.00 & 23,362 & 21.77 \\
\bottomrule
\end{tabular}
\caption{\textbf{Reasoning effort on tasks completed fully at every compared effort on OSWorld2.} Every reported score is 100\%. Time is in seconds per task; token rate divides total generated tokens by total agent time. Each model uses its own common task set.}
\label{tab:common-success}
\end{table}

\paragraph{Higher effort can generate longer outputs and more actions on the same successful tasks.}
Table~\ref{tab:common-success} restricts each OSWorld2 comparison to
tasks that receive full credit at every evaluated effort for that model.
On the five tasks that Gemini 3.8 Flash completes at all three efforts,
high effort takes 3,024\,s per task compared with 1,291\,s at low effort.
It generates more than twice as many tokens (73,578 versus 29,596) and
executes more actions and batches. In comparison, its effective token
rate falls from 47.57 to 42.18 tokens/s.

We observe a similar increase at Astra's highest effort. On the 13 tasks
it completes at all four settings, xhigh takes 1,132\,s, compared with
792\,s at low and 753\,s at medium. It generates 14,359 tokens and
executes 223 actions, compared with 7,878 tokens and 147 actions at low
effort. Muse Spark 1.3 completes two tasks at all five efforts, while
Kimi K3 has no task completed at all three efforts.

\clearpage
\section{Automated Search for Task-Selection Methods}
\label{app:autoresearch}

We investigate whether a coding agent can find task-selection methods
that better estimate a new model's score on the full task set. The agent
develops these methods using results from six models; we evaluate them
on a seventh model whose results are withheld throughout the search.

\begin{figure}[!t]
  \centering
  \includegraphics[width=\linewidth]{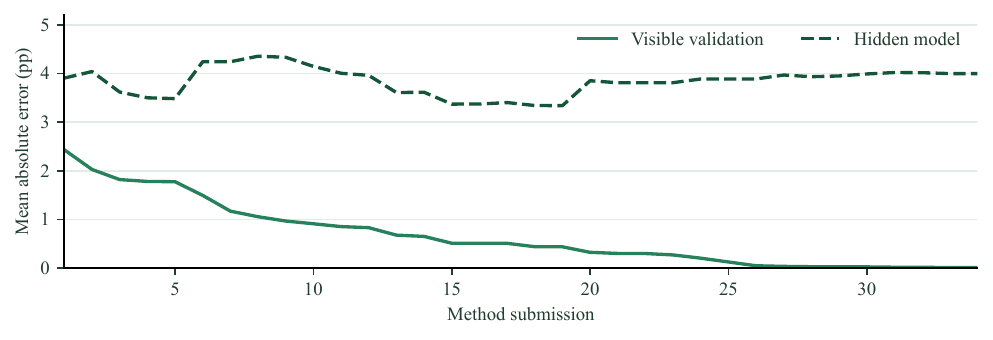}
  \caption{\textbf{Validation gains do not generalize to the held-out model.}
  Across 34 candidate methods, the coding agent receives only validation
  errors. We measure test error separately on the held-out model. Each method selects 32
  tasks. Error averages the mean absolute errors in partial score and
  exact completion, in percentage points.}
  \label{fig:autoresearch}
\end{figure}

\subsection{Search and Evaluation}

We provide the coding agent with results from six models on 107 OSWorld2
tasks. It can modify both how the tasks are selected and how scores on
the selected tasks are used to predict the full-set score. Each candidate
method selects exactly 32 tasks. We evaluate it using six-fold
leave-one-agent-out validation: five models are used to fit the method,
and the sixth to measure prediction error. The agent receives only this
validation error and can revise its method. We separately test each
candidate on the seventh model after fitting to all six available models.

\subsection{Validation Gains Do Not Generalize to the Held-Out Model}

Figure~\ref{fig:autoresearch} compares validation and test error across
34 candidate methods. Validation error falls from 2.43 to 0.01 percentage
points, while error on the held-out model changes from 3.90 to 4.00
percentage points. The search nearly eliminates validation error without
improving the estimate for the new model. Validation errors guide repeated
method selection, allowing the search to overfit these six models.

\stopcontents[appendices]

\end{document}